\documentclass[11pt,a4paper]{article}
\usepackage[utf8x]{inputenc}
\usepackage[pdftex]{graphicx} % Required for including pictures

\usepackage[%
    font={small,sf},
    labelfont=bf,
    format=hang,    
    format=plain,
    margin=0pt,
    width=0.8\textwidth,
]{caption}
\usepackage[list=true]{subcaption}

\usepackage[sort&compress,numbers,square]{natbib}
\usepackage{graphicx, color}

\usepackage{float}
\usepackage[pdftex,linkcolor=black,pdfborder={0 0 0}]{hyperref} % Format links for pdf

\usepackage{amsmath}
\usepackage{amsthm}
\usepackage{amssymb}

\title{The GREENBOT dataset: Multimodal mobile robotic dataset for a typical Mediterranean greenhouse}

\author{Fernando Ca\~{n}adas-Ar\'anega, Jose Luis Blanco-Claraco, \\Jose Carlos Moreno and Francisco Rodriguez}

\date{January 31st, 2024}

\begin{document}
	\maketitle

\begin{abstract}
This paper introduces an innovative dataset specifically crafted for challenging agricultural settings (a greenhouse), where achieving precise localization is of paramount importance. The dataset was gathered using a mobile platform equipped with a set of sensors typically used in mobile robots, as it was moved through all the corridors of a typical Mediterranean greenhouse featuring tomato crop. This dataset presents a unique opportunity for constructing detailed 3D models of plants in such indoor-like space, with potential applications such as robotized spraying. For the first time to the best knowledge of authors, a dataset suitable to put at test Simultaneous Localization and Mapping (SLAM) methods is presented in a greenhouse environment, which poses unique challenges. The suitability of the dataset for such goal is assessed by presenting SLAM results with state-of-the-art algorithms.
The dataset is available online in \url{https://arm.ual.es/arm-group/dataset-greenhouse-2024/}.
\end{abstract}

\section{Introduction}

In recent years, the use of technology in agriculture has experienced significant growth, driven by improved productivity and optimization of resources \citep{fan2012improving}. Technology is crucial in reducing the use of fertilizers, pesticides, and efficient water management \citep{evans2008methods}. Agricultural activities benefit considerably from technology, ranging from the application of micronutrients to the estimation of fruit quality and quantity, as well as the mechanical or electromagnetic removal of weeds \citep{bindraban2015revisiting}. Robots are presented as autonomous platforms with the ability to carry out these tasks and can also be used for example to collect a large amount of crop data continuously \cite{bechar2016agricultural}.

\begin{figure}
    \includegraphics[width=\linewidth]{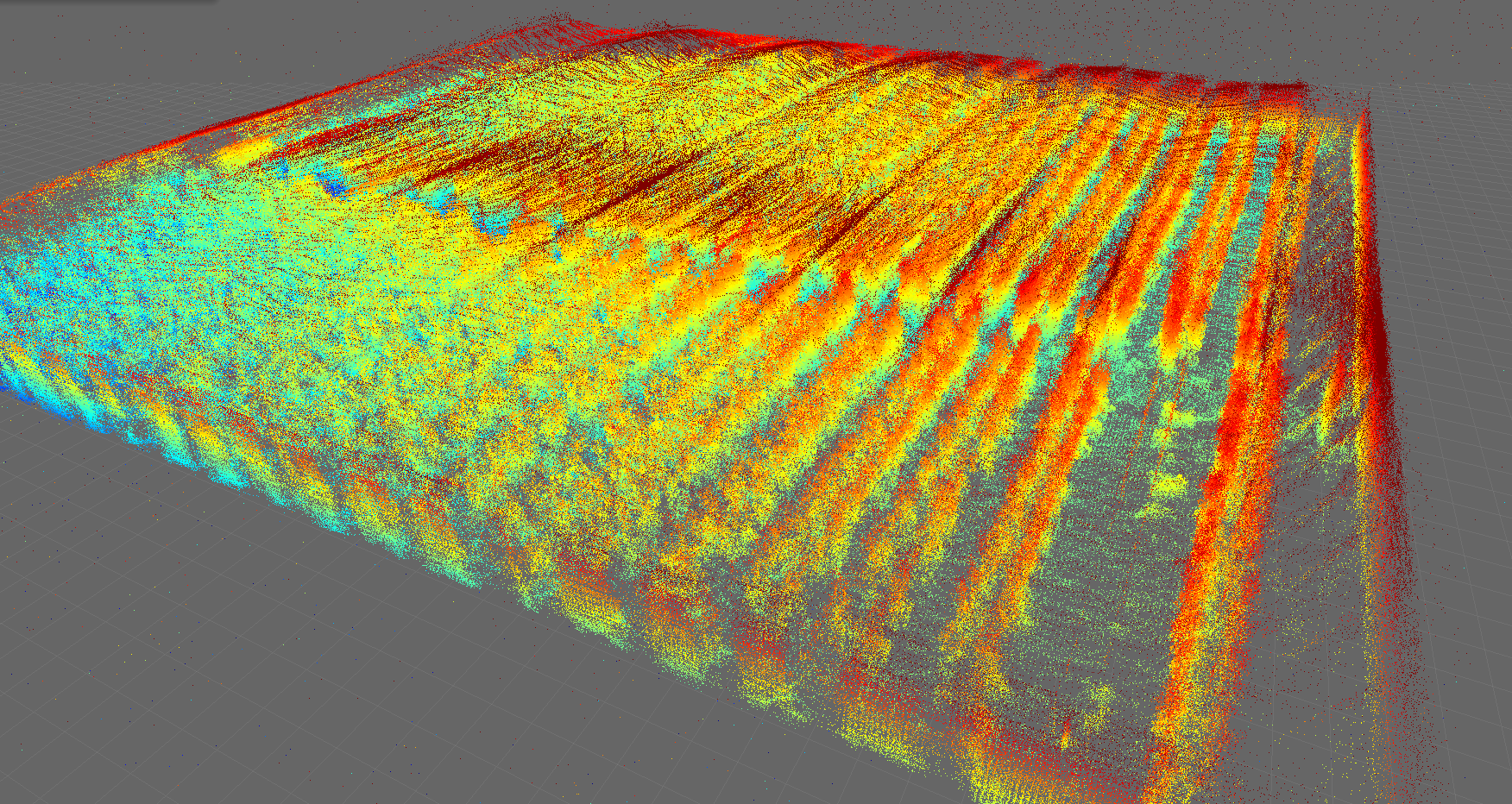} \centering
    \caption{Mapping of the complete greenhouse, orthogonal view with Velodyne VLP16. See discussion on section~\ref{sect:mapping.vlp16}.}
    \label{fig:Vel_SLAM2}
\end{figure}

Central to carrying out these tasks is providing a robot with autonomous navigation capabilities in an agricultural environment. In the particular case of a greenhouse, localization requires excellent accuracy and estimation of odometry, which is almost impossible to install in greenhouses \cite{yan2022real}. This environment presents particularly challenging features, such as repetitive patterns due to the lack of unique and consistent visual attributes of the vegetation, which poses unique challenges for closed-loop detection algorithms. In addition, these natural environments present unique computer vision challenges due to subtle human-caused movements, extreme light variability due to the closed environment, and seasonal crop changes due to crop management (pruning, harvesting,...) and weather variability.

Large and open datasets are crucial for developing data-driven solutions and benchmarks in a new era of deep learning-based algorithms. In the literature, there are abundant datasets for urban environments (e.g., \citep{geiger2013vision,maddern20171}) and, to the authors' knowledge, the most extensive datasets currently available for agricultural environments are the Rosario Dataset \citep{pire2019rosario}, the Sugar Beets Dataset \citep{chebrolu2017agricultural} and the MAGRO Dataset \citep{marzoa2023magro}, for an apple orchard field. However, these datasets are focused on open field agricultural environments where, on the one hand,  localization and orientation may be less complex and, on the other hand, the permissible error is greater than in closed environments such as greenhouses, in particular Mediterranean greenhouses, which accounts for 92\% of the total greenhouse area worldwide \citep{Rijswick2018,Hick2019}, characterized for being structured in irregular and narrow corridors (90-100 cm) with sandy soil (80\% of farms).

This paper presents an innovative and dynamic framework for data collection and a dataset collected throughout the growing season in a typical Mediterranean greenhouse using a sensor-equipped forklift. The data collected include images from a stereo camera pointing forward, an IMU and two 3D LIDARs. This dataset provides valuable information to advance the development of visual odometry. Notably, the data were acquired in a typical Mediterranean greenhouse with a tomato crop, which offers the opportunity to develop 3D models for the plants that can be specially useful, for example, for robotized spraying tasks, where the amount of plant protection product is applied while coordinating the relationship between the displacement speed of the robot and the spray speed \citep{Sanchez-Hermosilla2010, Sanchez-Hermosilla2013}. It is important to remark that the tomato is one of the most important crops in South-Eastern Spain, occupying 17\% of the greenhouse area in the province of Almería alone and accounting for 20\% of its total crop production during the 20/21 season \citep{CAGPDR2021}. 

From a robotics perspective, recent advancements have introduced several robotic solutions for agriculture automation. Most of these developments are not technologically mature, however some important companies in the agriculture machinery sector, such as John Deere, New Holland and Case, and other emerging companies with applications in this sector, such as Naïo and DJI, have commercial products or one step away from commercialization. However all of these developments have been conceived to work in open field. For greenhouses, robot prototypes performing different tasks such as weeding, harvesting, pruning, spraying, pest and disease monitoring, etc. have been also developed \citep{Kondo2011, Rodriguez2013}. Since 1987 many researchers have worked on the development of robots for greenhouses, studying the different problems involved \citep{Sanchez-Molina}. The AURORA project \cite{mandow1996autonomous}, for instance, introduced a resilient and cost-effective robot designed for greenhouse operations. This robot demonstrates the ability to navigate autonomously through diverse greenhouse environments. Another notable greenhouse automation initiative is \citep{Feng2018}, featuring a mobile robot equipped with a stereo vision system and a six-degree-of-freedom arm specifically designed for tomato cultivation in greenhouses. In addition, a service robot for sanitary control and localized dispensing of pesticides and fertilizers to plants in greenhouses using RGB-D cameras is presented in \cite{acaccia2003mobile}. Other projects in Japan \cite{kurata1994cultivation} have also received funding, contributing to the ongoing exploration and development of robotic solutions for greenhouse applications. It's important to notice that the authors of this paper have worked in three projects directly related to the use of mobile robots in greenhouses. In particular, \citep{GonzalezSnchez2009} presents the implementation and testing of navigation algorithms on the FITOROBOT mobile robot \citep{Sanchez-Hermosilla2010}. In INVERSOS project \cite{Sanchez-Hermosilla2013}, a multi-functional mobile robot is developed taking as basis the experience provided by the FITOROBOT project, and in the AGRICOBIOT \citep{Lopez-Gazquez2023} project another mobile robot for greenhouses is developed, but with the particular characteristic of working in a collaborative way close to humans.  

The need for this dataset is demonstrated by the fact that there are a multitude of robots that can improve the navigation task. The sensor-equipped platform used in this paper travels the same route in different conditions and on different days, capturing data with variations in lighting, weather, and crop state. Along with the sensors data, details on the calibration of intrinsic and extrinsic sensors parameters and tools for data processing are provided. 

The main contributions of this work are summarised as follows: (i) a novel dataset on an unexplored environment with rich and challenging features is presented; to the best knowledge of the authors, this is the first study on a dataset for mapping and navigation by mobile robots in greenhouses; (ii) an open-source data collection pipeline is developed and published, which will be used to augment and extend the dataset periodically, and (iii) the suitability of the proposed dataset is validated by running a state-of-the-art SLAM algorithm on it. The dataset, its download guide, and the tools to process it are publicly available in the dataset repository \footnote{University of Almería dataset "GREENBOT": \url{https://arm.ual.es/arm-group/dataset-greenhouse-2024/}}.

The rest of the paper is organized as follows. Section 2 presents a literature review, discussing the most relevant articles and focusing on works with application to agriculture. Section 3 presents the sensor-equipped platform and the sensors calibration. Section 4 presents how the data were acquired and the data structure. In Section 5, the dataset is evaluated with a novel algorithm to validate its quality, and finally, Section 6 ends with conclusions and future work.

\section{Related work}

Datasets tailored to mobile robotics are of great importance for tasks such as image-based localization, LIDAR odometry, and SLAM, among others.
Datasets can be classified according to whether they are synthetic (simulation of textures, objects, lighting, etc., from a virtual environment) or real (images or videos, point clouds, etc from a real environment), and according to the type of scene. 
Table \ref{tab:Dataset} summarizes the related datasets discussed in the rest of this section.

\subsection{Synthetic datasets}

Starting from the works related to rural environments, \cite{li2018interiornet} proposes a dataset based on a photorealistic environment where they evaluate many SLAM techniques, even going into interiors. In the same line, \cite{ROS2016synthia} develops a dataset based on urban photos, building an urban 3D environment and, following the same technique, \cite{giubilato2022challenges} focuses on mountainous environments. However, for works related to agricultural environments, \cite{yang2020multi} presents a synthetic dataset that evaluates different SLAM techniques with different types of plants in different camera positions, fruit labels, etc. In \cite{sinha2022high} a 3D environment of simulated vineyard data where it evaluates different mapping techniques is created. In \cite{lu2020survey}, authors make a synthetic dataset for different agricultural areas divided into different plots of land for spraying or pruning of fruit trees.

\begin{table}
\caption{Summary of the surveyed datasets}
\label{tab:Dataset}
\resizebox{\textwidth}{!}{
\begin{tabular}{ccccccllcccc}
\hline
\textbf{Dataset}                                 & \textbf{\begin{tabular}[c]{@{}c@{}}Real/ \\ synthetic\end{tabular}} & \textbf{\begin{tabular}[c]{@{}c@{}}Indoor/ \\ outdoor\end{tabular}} & \textbf{Environment}                                                & \textbf{\begin{tabular}[c]{@{}c@{}}Main \\ task\end{tabular}}                      & \textbf{RGB} & \multicolumn{1}{c}{\textbf{Depth}} & \multicolumn{1}{c}{\textbf{GPS}} & \textbf{IMU} & \textbf{LiDAR} & \textbf{\begin{tabular}[c]{@{}c@{}}Ground\\ truth\end{tabular}} & \textbf{\begin{tabular}[c]{@{}c@{}}Publicly \\ available\end{tabular}} \\ \hline
\cite{li2018interiornet}        & Synthetic               & Indoor                  & Room                                                                & SLAM                                                                               & Yes          & Yes                                & No                               & No           & No             & No                                                              & No                                                                     \\ \hline
\cite{ROS2016synthia}           & Synthetic               & Outdoor                 & Urban                                                               & \begin{tabular}[c]{@{}c@{}}Semantic \\ segmentation \\ for navigation\end{tabular} & Yes          & Yes                                & No                               & No           & No             & Pose                                                            & Yes                                                                    \\ \hline
\cite{giubilato2022challenges}  & Synthetic               & Outdoor                 & Mountain                                                            & SLAM                                                                               & No           & No                                 & Yes                              & Yes          & Yes            & Pose                                                            & Yes                                                                    \\ \hline
\cite{yang2020multi}            & Real                    & Outdoor                 & Open field                                                          & SLAM                                                                               & Yes          & Yes                                & No                               & No           & No             & \begin{tabular}[c]{@{}c@{}}Semantic\\ label\end{tabular}        & No                                                                     \\ \hline
\cite{lu2020survey}             & Synthetic               & Outdoor                 & \begin{tabular}[c]{@{}c@{}}Different \\ open field\end{tabular} & \begin{tabular}[c]{@{}c@{}}Semantic\\ segmentation\\ for recognise\end{tabular}    & Yes          & Yes                                & No                               & No           & No             & No                                                              & Yes                                                                    \\ \hline
\cite{maddern20171}             & Real                    & Outdoor                 & Urban                                                               & SLAM                                                                               & Yes          & No                                 & Yes                              & No           & Yes            & Pose                                                            & Yes                                                                    \\ \hline
\cite{geiger2013vision}         & Real                    & Outdoor                 & Urban                                                               & SLAM                                                                               & Yes          & Yes                                & Yes                              & Yes          & Yes            & Pose                                                            & Yes                                                                    \\ \hline
\cite{majdik2017zurich}         & Real                    & Outdoor                 & Urban                                                               & SLAM                                                                               & Yes          & No                                 & Yes                              & No           & Yes            & Pose                                                            & Yes                                                                    \\ \hline
\cite{blanco2014malaga}         & Real                    & Outdoor                 & Urban                                                               & SLAM                                                                               & No           & No                                 & Yes                              & Yes          & Yes            & Pose                                                            & Yes                                                                    \\ \hline
\cite{bandini2017measuring}     & Real                    & Outdoor                 & Lake and rivers                                                     & Measuring                                                                          & No           & \multicolumn{1}{c}{No}             & \multicolumn{1}{c}{Yes}          & Yes          & Yes            & Pose                                                            & No                                                                     \\ \hline
\cite{miller2018visual}         & Real                    & Outdoor                 & Rivers                                                              & SLAM                                                                               & Yes          & \multicolumn{1}{c}{No}             & \multicolumn{1}{c}{Yes}          & Yes          & No             & No                                                              & Yes                                                                    \\ \hline
\cite{de2021towards}            & Real                    & Outdoor                 & \begin{tabular}[c]{@{}c@{}}Various \\ open field\end{tabular}        & \begin{tabular}[c]{@{}c@{}}Semantic\\ segmentation\\ for navigation\end{tabular}   & Yes          & \multicolumn{1}{c}{Yes}            & \multicolumn{1}{c}{No}           & No           & No             & Pose                                                            & Yes                                                                    \\ \hline
\cite{liu2019pestnet}           & Real                    & Outdoor                 & Open field                                                          & Pest detection           & Yes          & \multicolumn{1}{c}{Yes}            & \multicolumn{1}{c}{No}           & No           & No             & No                                                              & No                                                                     \\ \hline
\cite{pire2019rosario}      & Real                                                               & Outdoor                                                            & Soybean field                                                                        & SLAM                                                           & Yes          & Yes                                & Yes                              & Yes          & No             & Pose                                                            & Yes                                                                    \\ \hline
\cite{chebrolu2017agricultural} & Real                    & Outdoor                 & Sugar beet field                                                    & Weeding               & Yes          & \multicolumn{1}{c}{Yes}            & \multicolumn{1}{c}{Yes}          & Yes          & Yes            & Pose                                                            & Yes                                                                    \\ \hline
\cite{marzoa2023magro}      & Real                                                               & Outdoor                                                            & Apple field                                                                          & SLAM                                                           & Yes          & Yes                                & Yes                              & No          & Yes            & Pose                                                            & Yes                                                                    \\ \hline

\cite{karam2022microdrone}      & Real                                                               & Outdoor                                                            &  Building                                                                          & SLAM                                                           & No          & No                                & Yes                              & Yes          & Yes            & Pose                                                            & Yes                                                                    \\ \hline

\cite{kirsanov2019discoman} & Real                                                               & Indoor                                                             & House                                                                                & SLAM                                                           & Yes          & No                                 & No                               & No           & No             & Pose                                                            & No                                                                     \\ \hline
\cite{lee2018monocular}     & Real                                                               & Indoor                                                             & Laboratory                                                                           & SLAM                                                           & Yes          & Yes                                & No                               & No           & No             & Pose                                                            & Yes                                                                    \\ \hline

\cite{martin2021generic}    & Real                                                               & Outdoor                                                            & \begin{tabular}[c]{@{}c@{}}Laboratory scale \\ greenhouse\end{tabular}  & \begin{tabular}[c]{@{}c@{}}Navigation\\ techniques\end{tabular} & Yes          & No                                 & No                               & No           & No             & No                                                              & No                                                                     \\ \hline

\cite{brostow2009semantic}      & Real                                                               & \begin{tabular}[c]{@{}c@{}}Indoor/\\ Outdoor\end{tabular}          & \begin{tabular}[c]{@{}c@{}}Mountain and \\ building\end{tabular}                     & SLAM                                                           & Yes          & No                                 & Yes                              & Yes          & No             & Pose                                                            & No                                                                     \\ \hline

GREENBOT      & Real                                                               & Indoor         & \begin{tabular}[c]{@{}c@{}}Mediterranean \\ greenhouse\end{tabular}                     & SLAM                                                           & Yes          & Yes                                 & No                              & Yes          & Yes             & Pose                                                            & Yes                                                                     \\ \hline
\end{tabular}
}
\end{table}

\subsection{Real datasets}

Within these real datasets, scenes can be classified as indoor or outdoor, with outdoor scenes classifiable according to whether they are recording urban locations or other types of environments.

\subsubsection{Outdoors}

In terms of environments, outdoor areas often present more difficulties due to changes in lighting conditions, weather, moving objects, etc. These can be further classified into urban or open field environments, depending on the environment where the data were collected. The latter is particularly interesting for the present work because, although it is carried out in a greenhouse, which can be classified as indoor in agriculture, it is crucial to compare it with other open field agriculture datasets, highlighting the non-existence of datasets for indoor agriculture environments such as Mediterranean greenhouses.

\begin{itemize}
    \item \textbf{Urban environments}
\end{itemize}

In \cite{maddern20171} authors present the dataset for a university campus where data are collected using six RGB cameras, a LiDAR, and a GPS. In \cite{geiger2013vision} high-resolution color and greyscale stereo cameras, a Velodyne 3D laser scanner, and a high-precision GPS/IMU are used to collect data from a city. These studies collected data using ground vehicles. In \cite{majdik2017zurich}, data were captured by flying at low altitudes within urban streets using an aerial micro-vehicle equipped with high-resolution cameras, GPS, and IMU. A dataset based on a high-resolution stereo camera at 20~Hz for 36.8~km in Málaga with a Citroen C4 car is presented in \cite{blanco2014malaga}.

\begin{itemize}
    \item \textbf{Open field environments}
\end{itemize}

With the help of human-operated vehicles, \cite{bandini2017measuring} collected data of the river and lake vicinity over three years equipped with a 2D LiDAR, RGB camera, GPS, and IMU. Similarly, \cite{miller2018visual} managed to gather data from the Sangamon River using a canoe equipped with an RGB camera, IMU, and GPS. Focusing on agriculture in open fields, \cite{de2021towards} classified the data obtained through an agricultural robot in a beet plantation for three months, acquiring data from an RGB-D camera, IMU, GPS, and LiDAR. Moreover, there are also articles addressing agricultural tasks unrelated to odometry, mapping, or navigation. In \cite{liu2019pestnet}, a dataset for pest or harmful insect detection is presented, as well as fruit and plant detection \cite{duggal2016plantation}, among others. The related existing works with the most impact in agriculture focus on the dataset collected in Rosario \cite{pire2019rosario} and in the Sugar Beets \cite{chebrolu2017agricultural}, where, with the help of mobile robots, they can map by blending techniques based on georeferenced images and SLAM algorithms.  These works are focused on weed detection, mapping of  different types of crops, etc. Finally, in \cite{marzoa2023magro}, a dataset is presented using a robot with autonomous navigation, displaying images with RGB-D cameras and 3D LiDAR on a row of apple trees, closing the control loop through a SLAM algorithm.

\subsubsection{Indoors}

Although most datasets focus their work on outdoors, numerous studies on indoors datasets exist. In this work, after conducting an exhaustive search, they will be divided into buildings and greenhouses. However, as mentioned above, few works discuss datasets within greenhouses. In particular for Mediterranean greenhouses, to the authors' knowledge, there is none. 

\begin{itemize}
    \item \textbf{Building environments}
\end{itemize}

Depending on the scheme to be recorded, these datasets can adopt different types of complexity. If the data are recorded in indoor environments such as rooms, they are usually easier to recompile and process as they do not experience changes in lighting or significant differences related to seasonal changes. In addition, these environments are fabricated by objects that the user can easily place. In this case, there are a multitude of datasets that recompile authentic interiors as, for example, for a regular house,  in \cite{kirsanov2019discoman}, a SLAM technique based on RGB cameras using different robotic platforms is presented; in \cite{karam2022microdrone}, a drone equipped with a stereo camera and an IMU acquires data from the interior of an industrial building. However, the variation of light, natural light, or the need to know the whole environment can lead to a disorientation of the robot, producing an erroneous map. In this case, in \cite{lee2018monocular} authors contemplate the different casuistics that a service robot can experience when it navigates in interiors.

\begin{itemize}
    \item \textbf{Agriculture environments}
\end{itemize} 

Localization by means of GNSS is fundamental for navigation with mobile robots \cite{wang2018statistical}, although weak signal reception inside greenhouses leads to many methods exploiting alternatives 
such as rail-based location \cite{ko2014autonomous}, or beacons, such as QR codes \cite{feng2018design}. To date, most of the work on robots in greenhouses has been focused on (i) weed inspection and spraying and (ii) vegetable harvesting tasks. Regarding datasets, in \cite{martin2021generic}, a small ROS dataset is presented for data collection and mapping using different techniques with RGB cameras, with an unpublished proprietary dataset. Similarly, semantic mapping is performed in \cite{matsuzaki20183d} to address contextual navigation in a laboratory scale greenhouse with agricultural robots, using the CamVid dataset \cite{brostow2009semantic}. Other recent datasets focus on detailed point clouds for phenotyping plants \cite{marks2023bonnbeetclouds3d}.

These works do not provide a detailed dataset that can be used to evaluate different SLAM techniques. The proposed work is entirely novel as there is no dataset using two widely used LiDAR models and cameras navigating through different crop rows inside a greenhouse.

\section{Platform description}

The platform used to record this dataset was designed at the University of Almería \footnote{With the support of the Regional Ministry of Economic Transformation, Industry, Knowledge and Universities and the European Regional Development Fund (FEDER) with the projects UAL2020-TEP-A1991 and PY2020\_007-A1991.} (Figure \ref{fig:Calibration}). It has a base sized 900 x 500 x 700 mm with a load capacity of 180 kg, equipped with four wheels and two handles for manual operation. The sensors are mounted on top of it, taking into account the specific constraints of each sensor:

\begin{itemize}
    \item \textit{Stereo camera} \footnote{Bumbleblee BB2-08S2 (\href{https://www.flir.com/support/products/bumblebee2-firewire/}{website}).}: It has two lenses for capturing synchronized stereo images. It connects to the PC through a FireWire IEEE 1394 interface, providing a data transmission speed of 800 Mb/s. The stored data has a resolution of 1032 x 776 pixels. The field of view is 97$^\circ$ horizontally and 66$^\circ$ vertically within a range of 0.3 m to 20 m. Recording was done at 10 Hz, with a maximum frame rate of 20 fps. This camera is placed on the front so that the images captured are ideal for obstacle identification for a robot.

    \item \textit{Velodyne VLP16 LiDAR} \footnote{Velodyne VLP16 (\href{https://velodynelidar.com/products/puck/}{website}).}: Velodyne's LiDAR VLP16 with a maximum range of 100 m, a 360$^\circ$ horizontal and 30$^\circ$ vertical field of view is used. It is placed at the top of the runway to have the whole field of view available, except for the operator. It was recorded with a frequency of 10 Hz.

    \item \textit{Ouster OS0 LiDAR} \footnote{Ouster OS0 (\href{https://ouster.com/products/hardware/os0-lidar-sensor}{website}).}: Data are also collected from the Ouster model OS0. This is a 32-channel LiDAR with 360$^\circ$ horizontal and 90$^\circ$ vertical field of view, ranging between 0.3 m and 50 m. The data rate is also left at the default value of 10 Hz, including the onboard 6-axis IMU. This sensor is placed in a structure 20 cm lower than the Velodyne, obstructing its view and hence only collecting data with an effective horizontal field of view of 275$^\circ$.
    
\end{itemize}

A NovAtel GNSS, model IMU-IGM-A1 \footnote{NovAtel (\href{https://novatel.com/support/span-gnss-inertial-navigation-systems/span-combined-systems/span-igm-a1}{website}).}, is also installed with the antenna ANTCOM 42G1215 \footnote{ANTCOM (\href{https://novatel.com/products/gps-gnss-antennas/compact-small-gnss-antennas/42g1215a-xt-1}{website}).}.
In this dataset, GNSS was only used for timestamp synchronization.  

\subsection{Computing platform}

Greenhouses often have hard environmental conditions, with high temperature and humidity
\cite{hernandez2017microclimate}. Additionally, power for sensors and computer must be supplied from a battery that balances versatility and space. Taking all this into account, the platform is equipped with a HISTTON computer featuring an Intel i7 - 8550U processor (4 GHz), an Intel UHD 620 graphics card with 24 CUDA cores, and 32 GB of DDR4 RAM. This equipment was selected for its operating temperature range of 0 - 70 ºC and humidity range between 0 and 85\%. It only consumes 15 W, resulting in minimal power consumption to consider for battery capacity. This computer was equipped with two disk partitions to use both, Ubuntu 20.04 with ROS Noetic, and Ubuntu 22.04 with ROS~2 Humble, providing the opportunity to work with both versions.

\subsection{Calibration}

\begin{figure}
    \includegraphics[width=12cm]{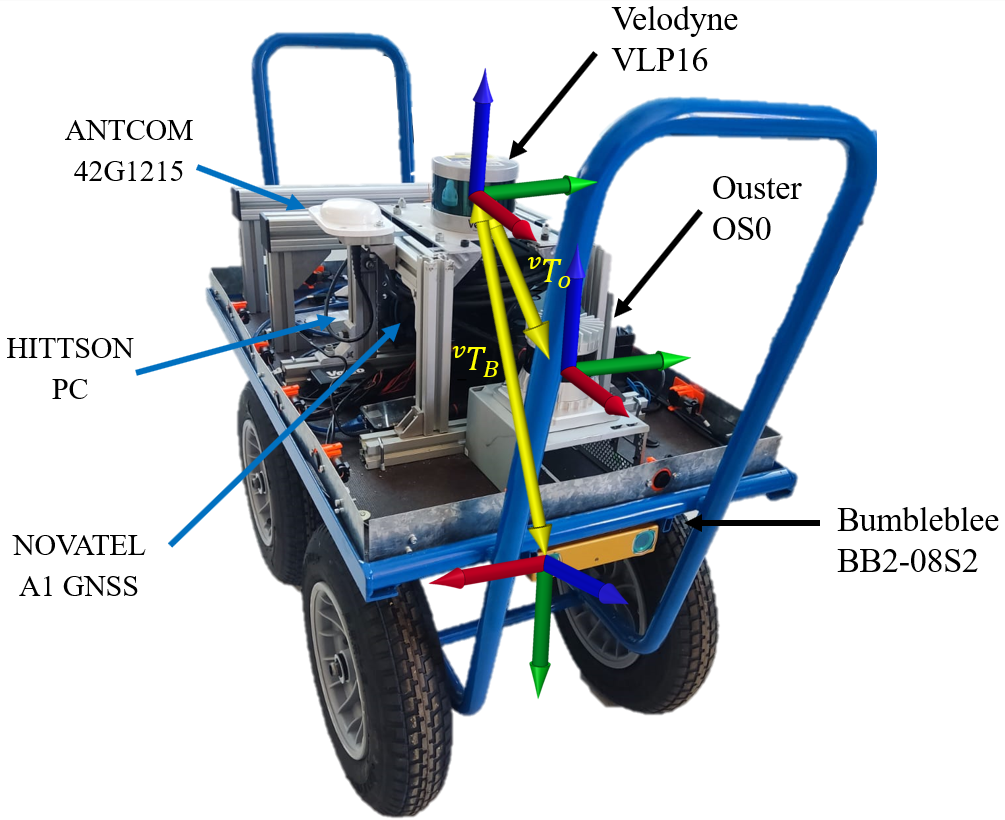}\centering
    \caption{Sensor-equipped platform used to acquire data, equipped with a Bumblebee stereo camera, Novatel GNSS, ANTCOM 45G1215, a Velodyne VLP16 3D LiDAR, and an Ouster OS0 LiDAR with IMU. Also, map of reference local systems for platform sensors. The x-axis is represented in red, the y-axis in green, and the z-axis in blue.}
    \label{fig:Calibration}
\end{figure}

\subsubsection{Intrinsic parameters}

\begin{figure}[H]
    \includegraphics[width=\linewidth]{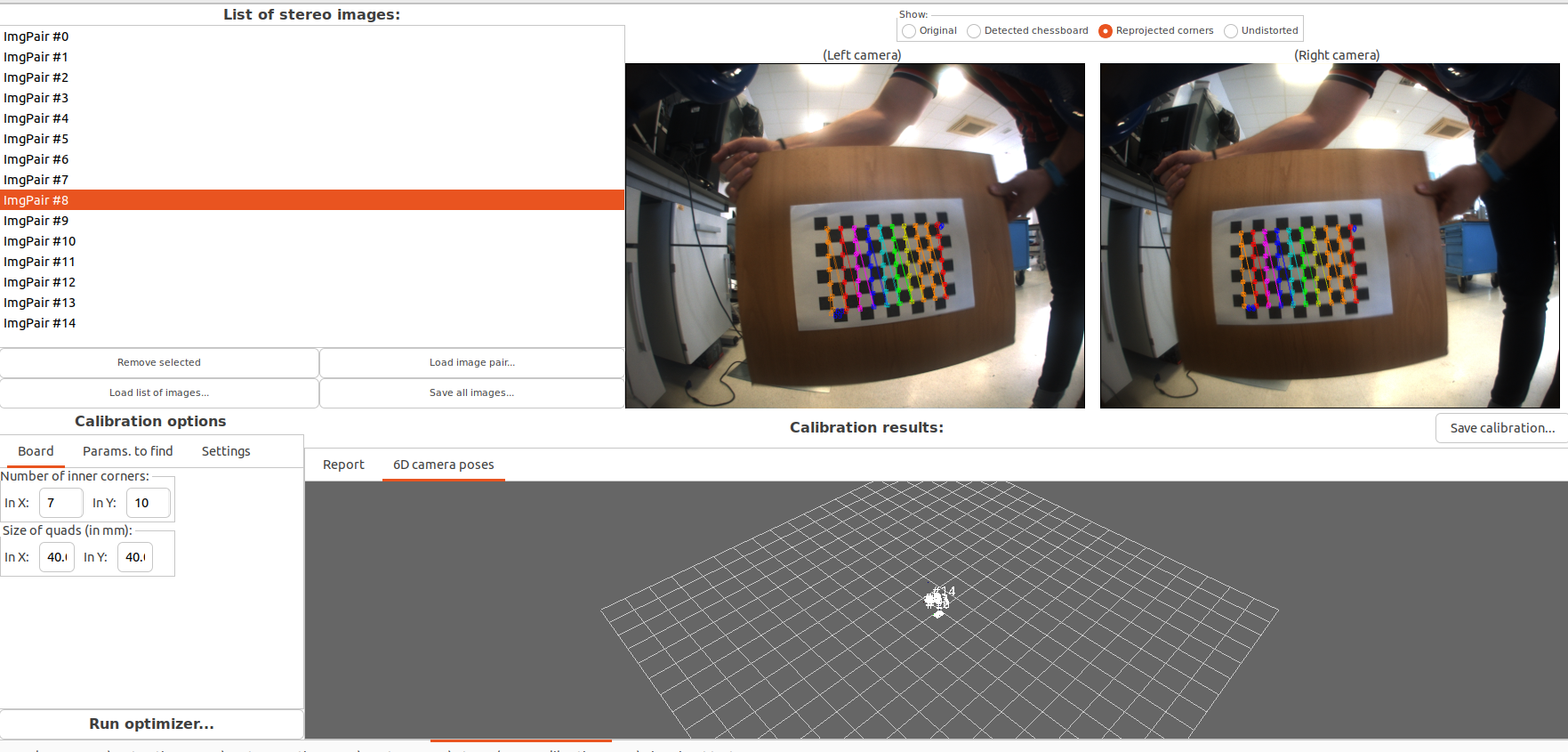}\centering
    \caption{Screenshot of the intrinsic calibration application used for the stereo camera \cite{mrpt2024}.}
    \label{fig:Calib}
\end{figure}

The intrinsics of the stereo camera have been determined with the \texttt{kinect-stereo-calib} application from 
the Mobile Robot Programming Toolkit (MRPT) \cite{mrpt2024}, including the focal lengths $(fx,fy)$, the optical centers $(cx,xy)$, the pin-hole distortion parameters $(k1,k2,p1,p2)$, and the left-to-right relative pose.
Calibration was done using a 7x10 classic checkerboard target pattern, capturing 20 pairs of images.
The calibration file is available online.
An example of the calibration tool, showing the result, is shown in Figure~\ref{fig:Calib}.

\subsubsection{Extrinsic parameters}

Extrinsic parameters define the relative poses of the different sensor frames of reference. The origin of coordinates is defined at the Velodyne, from which the other sensor poses are defined, as shown in Figure \ref{fig:Calibration}:

\begin{itemize}
    \item Stereo camera: x = right, y = down, z = forward.
    \item LIDARs: x = forward, y = left, z = up.
\end{itemize}

The Velodyne sensor is taken as a local frame of reference for the platform. 

\section{GREENBOT dataset}

This section describes the environment, data acquisition procedure, download method, and usage instructions.

\subsection{Greenhouse and crop}

\begin{figure}[H]
    \includegraphics[width=\linewidth]{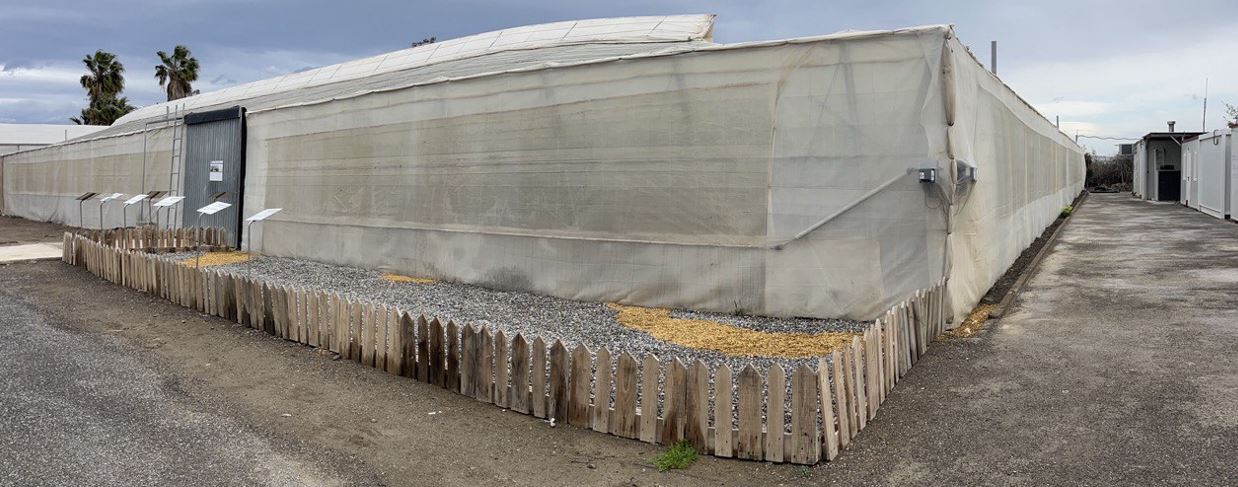} \centering
    \caption{IFAPA experimental greenhouse.}
    \label{fig:Grennhouse}
\end{figure}

The experiments took place at the Agroconnect facilities (which has received cofunding from the Ministry of Science, Innovation, and Universities in collaboration with the European Regional Development Fund (FEDER) under the grant program for acquiring cutting-edge scientific and technological equipment (2019)) in the Municipal District of La Cañada de San Urbano, Almería, located at 36°50' N and 2°24' W, with an elevation of 3 meters above sea level and a slope in the terrain of 1 \% in the North direction (see Figure \ref{fig:Grennhouse}).

The greenhouse is in one of the most common styles in the region (Almería's "raspa y amagado"),
expands over 1850 square meters, and features a sturdy steel frame and a polyethylene covering. The greenhouse is arranged in an East-West ridge configuration to benefit
from the natural ventilation from those two predominant wind directions in the region.
A 2-meter wide central pathway in the greenhouse serves as the main road and leads to eleven aisles on each side. The aisles on the North side are 2 meters wide and 12.5 meters long, while the aisles on the South side are 2 meters wide and 22.5 meters long. Radiating from this central aisle are narrower secondary pathways, each only one meter wide, facilitating the smooth movement of mobile robotic units. The facility has two sections with various advanced systems tailored for precise crop management. These systems include natural ventilation from above (zenithal) and along the sides (lateral), an integrated air heating and cooling network, infrastructure for CO2 enrichment, cutting-edge equipment for humidification and dehumidification, an advanced irrigation and fertilization system, as well as energy-efficient LED artificial lighting. 

\begin{figure}[H]
    \includegraphics[width=5.45cm]{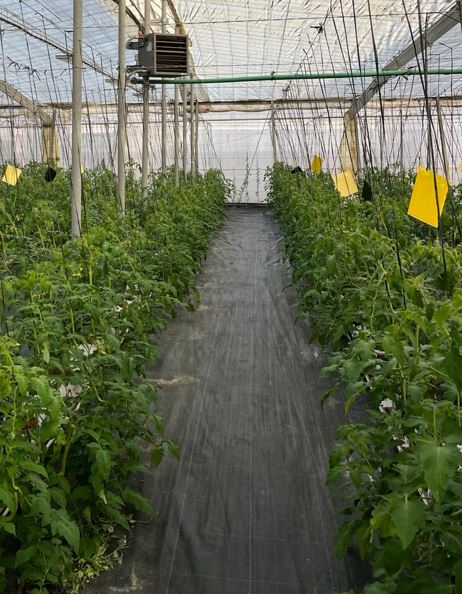} \centering
    \includegraphics[width=7cm]{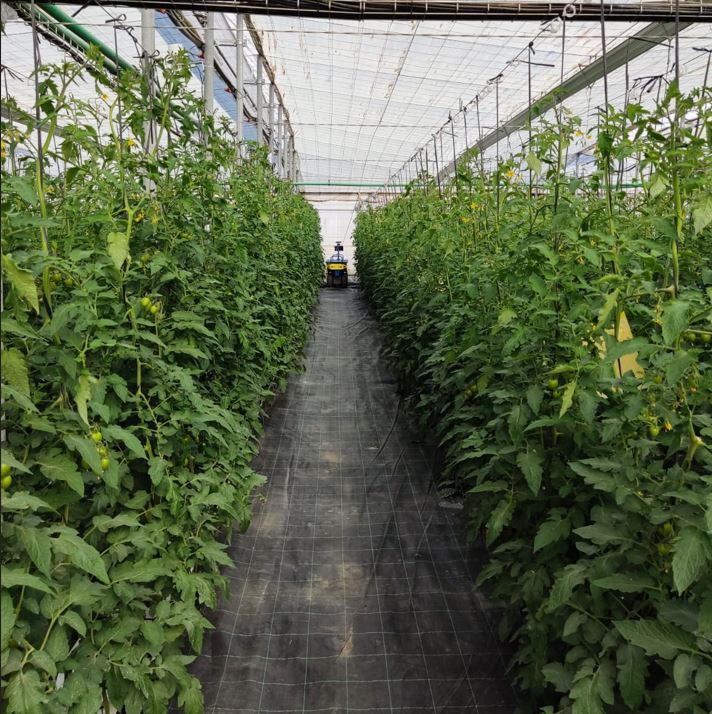} \centering
    \caption{Example of two images acquired at the same position with different crop stages: the first one (left) after one month of plantation and the second one (right) after three months}
    \label{fig:Cultivo}
\end{figure}

The greenhouse crop is tomato (\textit{Lycopersicon esculentum}), and the plants are grown in coir bags, in rows oriented from north to south with a slope of 1\%, as shown in Figure \ref{fig:Cultivo}.

\subsection{Data acquisition}

\begin{figure}[H]
    \includegraphics[width=\linewidth]{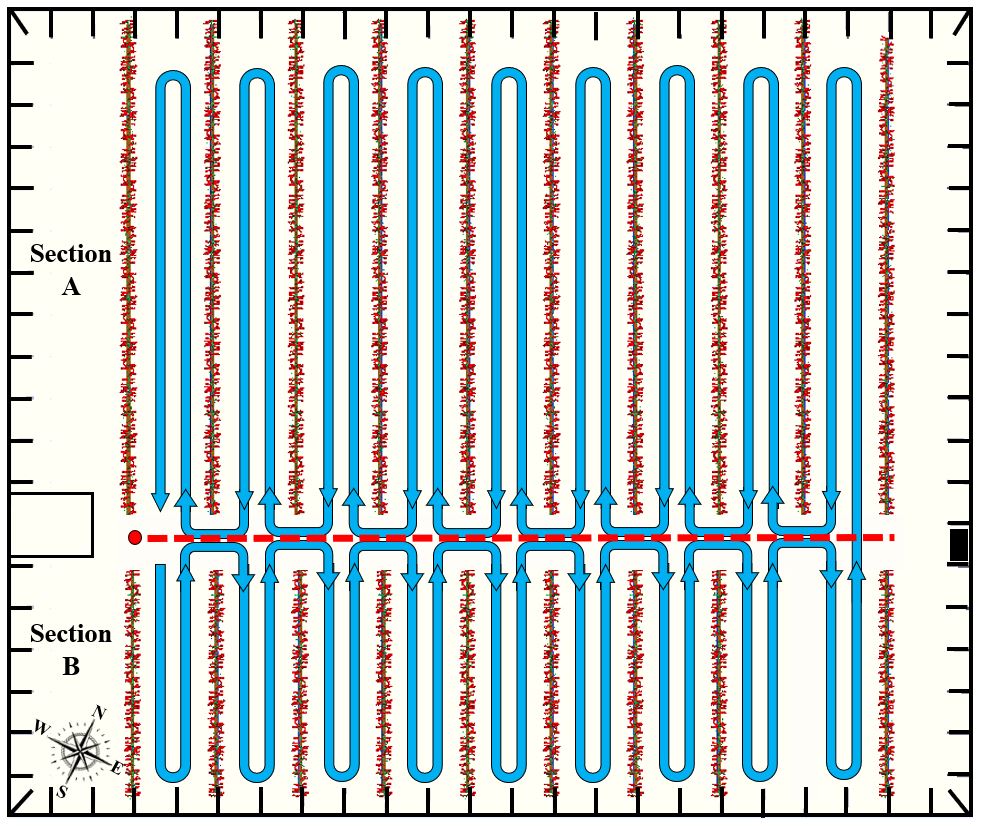} \centering
    \caption{Trajectory taken during the data collection campaign in the greenhouse of the IFAPA centre, Almería. The red dot indicates the start and end of the route.}
    \label{fig:Tarjectoria}
\end{figure}

The trajectory that has been followed through the crop is shown in Figure \ref{fig:Tarjectoria}. In order to distinguish the longer corridors from the shorter ones, the greenhouse is segmented into two sections, dividing it by the central corridor.

The platform is driven by the operator at an average velocity of 0.83 [m/s], who makes one pass until the end of each aisle and another pass on the way back, then switching to the next row. It should be noted that the ninth row in section B did not have plants at the time of recording this dataset. In order to carry out this data collection, the point cloud provided by the Velodyne and the Ouster LIDARs, the Ouster IMU, and the stereo images from the Bumblebee camera are stored in ROS 1 bag format. 
In the aisles located to the east and west, it has not been possible to make passes due to the impossibility of passing through the greenhouse's supporting pillars.

In terms of data structure, there are a total of \emph{nine sequences}, recorded between October 5th, 2022 and December, 1st 2022.
Weekly recordings were made, subjected to different environmental conditions, lighting, ground changes, etc. The first two data files contain information from section B only (refere to Figure \ref{fig:Tarjectoria}, as section A was temporarily inoperative).
The third sequence contains section A, separated from section B. Finally, the remaining datasets contain a single \textit{rosbag} that stores the entire greenhouse. Table \ref{tab:Datos} shows detailed and summarised information about the stored data, providing information about the weather on the day of recording and the height of a pilot plant measured manually.
The path length of each sequence is obtained from 
the SLAM-based reconstructions detailed later on,
using the \texttt{evo\_traj} tool \cite{grupp2017evo}.
Finally, the climatological variables of air temperature, humidity, and irradiance inside the greenhouse during the navigation, are extracted from a database of the University of Almería and also incorporated to the dataset.

\begin{table}[H]
\caption{Description of each recorded segment}
\label{tab:Datos}
\resizebox{\textwidth}{!}{
\begin{tabular}{cccccc}
\hline
\textbf{Sequence} & \textbf{\begin{tabular}[c]{@{}c@{}}Length\\  {[}m{]}\end{tabular}} & \textbf{\begin{tabular}[c]{@{}c@{}}Duration\\  {[}s{]}\end{tabular}} & \textbf{Section} & \textbf{\begin{tabular}[c]{@{}c@{}}Greenhouse\\ condition\end{tabular}}                                                      & \textbf{Description}                                                                  \\ \hline
\texttt{2022\_10\_05}      & 459.25                                                                & 696                                                                  & B                & \begin{tabular}[c]{@{}c@{}}Temp: 27.01 {[}ºC{]}\\ Hum: 62.72 {[}\%{]}\\ Ir: 106.7 {[}W/m\textasciicircum{}2{]}\end{tabular} & \begin{tabular}[c]{@{}c@{}}Sunny, morning\\ plant height\\ 0.96 {[}m{]}\end{tabular}  \\ \hline
\texttt{2022\_10\_14}      & 457.36                                                                & 701                                                                  & B                & \begin{tabular}[c]{@{}c@{}}Temp: 23.78 {[}ºC{]}\\ Hum: 63.12 {[}\%{]}\\ Ir: 79.0 {[}W/m\textasciicircum{}2{]}\end{tabular}  & \begin{tabular}[c]{@{}c@{}}Cloudy, morning\\ plant height\\ 1.05 {[}m{]}\end{tabular} \\ \hline
\texttt{2022\_10\_19}      & 1321.21                                                                 & 1432                                                                 & A \& B            & \begin{tabular}[c]{@{}c@{}}Temp: 25.21 {[}ºC{]}\\ Hum: 75.83 {[}\%{]}\\ Ir: 87.9 {[}W/m\textasciicircum{}2{]}\end{tabular}  & \begin{tabular}[c]{@{}c@{}}Sunny, morning\\ plant height\\ 1.12 {[}m{]}\end{tabular}  \\ \hline
\texttt{2022\_10\_26}      & 1432.08                                                                & 1463                                                                 & A \& B            & \begin{tabular}[c]{@{}c@{}}Temp: 23.22 {[}ºC{]}\\ Hum: 60.09 {[}\%{]}\\ Ir: 66.4 {[}W/m\textasciicircum{}2{]}\end{tabular}  & \begin{tabular}[c]{@{}c@{}}Cloudy, morning\\ plant height\\ 1.35 {[}m{]}\end{tabular} \\ \hline
\texttt{2022\_11\_02}      & 1233.87                                                               & 1486                                                                 & A \& B            & \begin{tabular}[c]{@{}c@{}}Temp: 15.84 {[}ºC{]}\\ Hum: 72.35 {[}\%{]}\\ Ir: 70 .7 {[}W/m\textasciicircum{}2{]}\end{tabular} & \begin{tabular}[c]{@{}c@{}}Cloudy, morning\\ plant height\\ 1.41 {[}m{]}\end{tabular} \\ \hline
\texttt{2022\_11\_09}      & 1293.29                                                                 & 1532                                                                 & A \& B            & \begin{tabular}[c]{@{}c@{}}Temp: 16.23{[}ºC{]}\\ Hum: 62.7 {[}\%{]}\\ Ir: 82.7 {[}W/m\textasciicircum{}2{]}\end{tabular}    & \begin{tabular}[c]{@{}c@{}}Cloudy, morning\\ plant height\\ 1.53 {[}m{]}\end{tabular} \\ \hline
\texttt{2022\_11\_19}      & 1332.58                                                                 & 1752                                                                 & A \& B            & \begin{tabular}[c]{@{}c@{}}Temp: 17.45 {[}ºC{]}\\ Hum: 62.7 {[}\%{]}\\ Ir: 64.36 {[}W/m\textasciicircum{}2{]}\end{tabular}  & \begin{tabular}[c]{@{}c@{}}Sunny, morning\\ plant height\\ 1.60 {[}m{]}\end{tabular}  \\ \hline
\texttt{2022\_11\_23}      & 1428.45                                                                & 1692                                                                 & A \& B            & \begin{tabular}[c]{@{}c@{}}Temp: 15.28 {[}ºC{]}\\ Hum: 62.7 {[}\%{]}\\ Ir: 69.4 {[}W/m\textasciicircum{}2{]}\end{tabular}   & \begin{tabular}[c]{@{}c@{}}Cloudy, morning\\ plant height\\ 1.73 {[}m{]}\end{tabular} \\ \hline
\texttt{2022\_11\_30}      & 1440.32                                                               & 1730                                                                 & A \& B            & \begin{tabular}[c]{@{}c@{}}Temp: 16.05 {[}ºC{]}\\ Hum: 62.7 {[}\%{]}\\ Ir: 74.4 {[}W/m\textasciicircum{}2{]}\end{tabular}   & \begin{tabular}[c]{@{}c@{}}Cloudy, morning\\ plant height\\ 1.85 {[}m{]}\end{tabular} \\ \hline
\end{tabular}
}
\end{table}

\subsection{Data structure}

As already mentioned, the data are stored in \textit{rosbag} format as it allows storing the desired topics in the original format. 
This rosbag keeps the ranging data from both LIDARs, 
the stereo images, and the IMU from the Ouster LIDAR.

Alternatively, LIDAR data are also provided in the 
\textit{rawlog} format\footnote{Description of the \textit{rawlog} format \href{https://docs.mrpt.org/reference/latest/robotics_file_formats.html}{available online}.}.
This format can be easily parsed and published by ROS~2 packages such as \texttt{mrpt\_rawlog} or the SLAM framework described later on.

\section{SLAM suitability assesment}

This section will assess the applicability of the dataset
to build maps using a LIDAR SLAM framework. 
In particular, we used \texttt{mola\_lidar\_odometry} from the MOLA framework \cite{MOLA}.
The approach belongs to the family of LIDAR odometry methods (i.e. without loop closure) with voxel-based 
raw point cloud representation.
Experimental results from both LiDARs are presented separately in the following subsections.

\subsection{Mapping - Velodyne VLP16}
\label{sect:mapping.vlp16}

\begin{figure}[H]
  \centering
    \subcaptionbox{}{\includegraphics[width=0.47\textwidth]{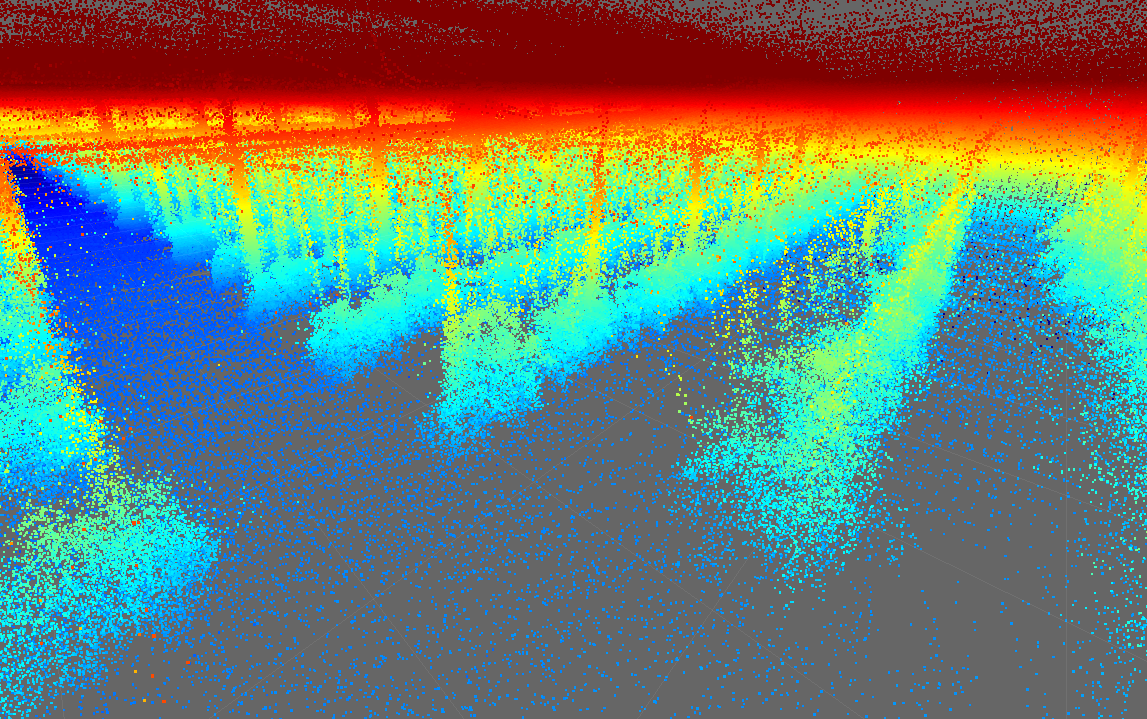}}%
    \label{subfig:2022-10-14}
  \hfill
    \subcaptionbox{}{\includegraphics[width=0.47\textwidth]{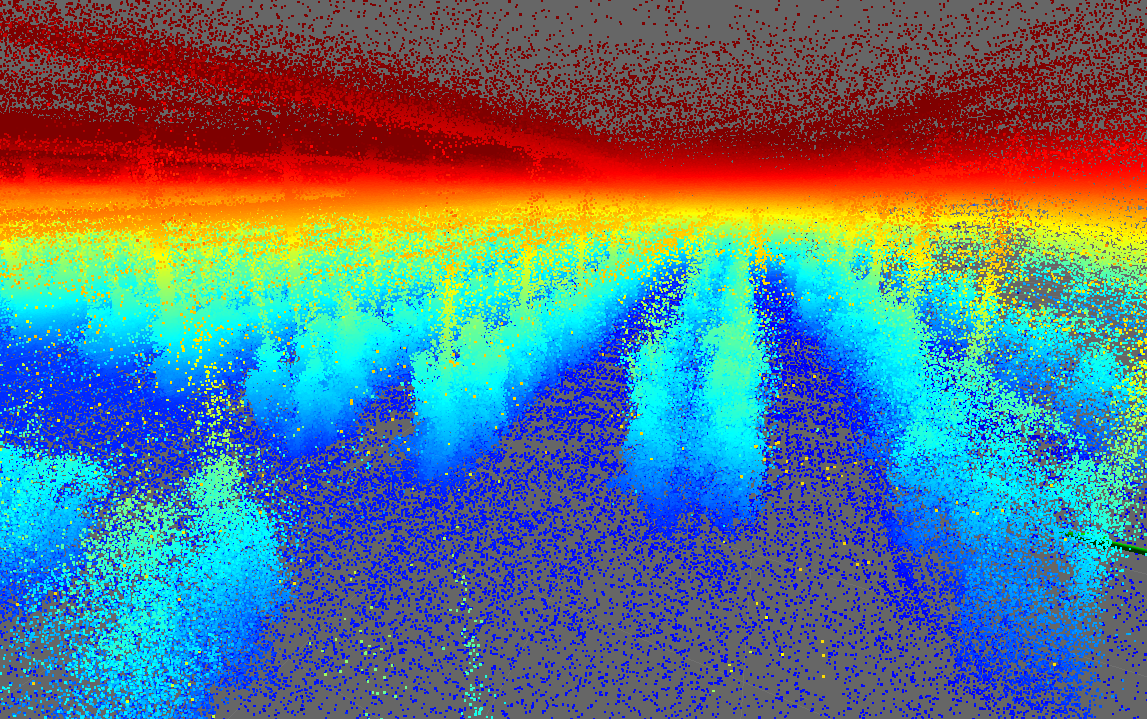}}%
    \label{subfig:2022-10-26}
    \\
    \subcaptionbox{}{\includegraphics[width=0.47\textwidth]{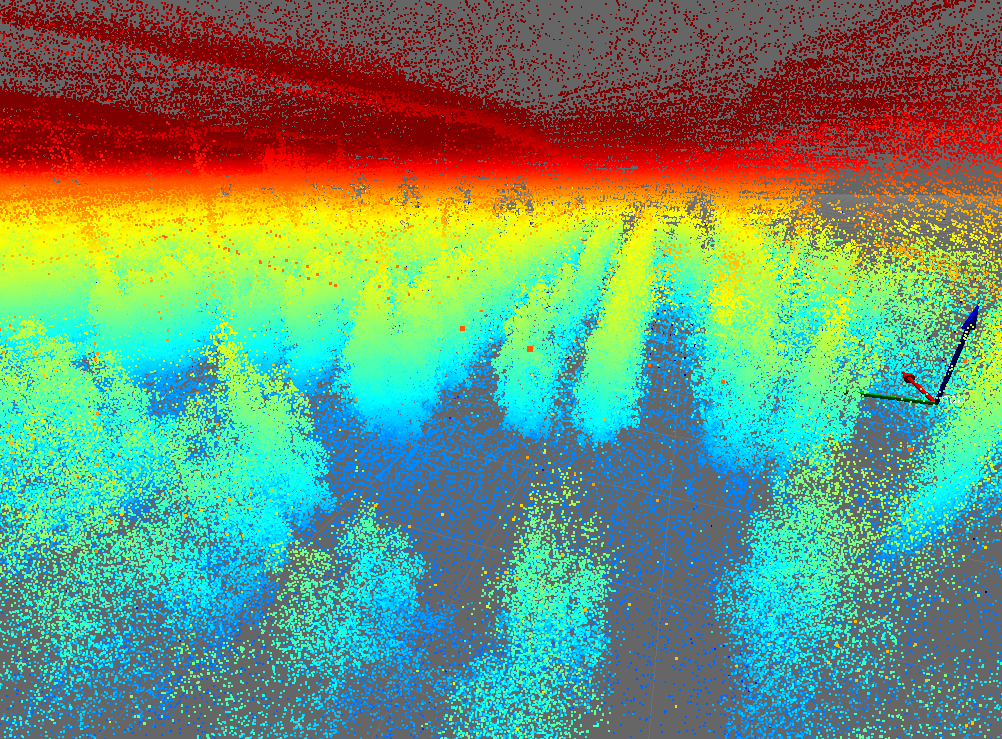}}%
    \label{subfig:2022-11-09}
  \hfill
    \subcaptionbox{}{\includegraphics[width=0.47\textwidth]{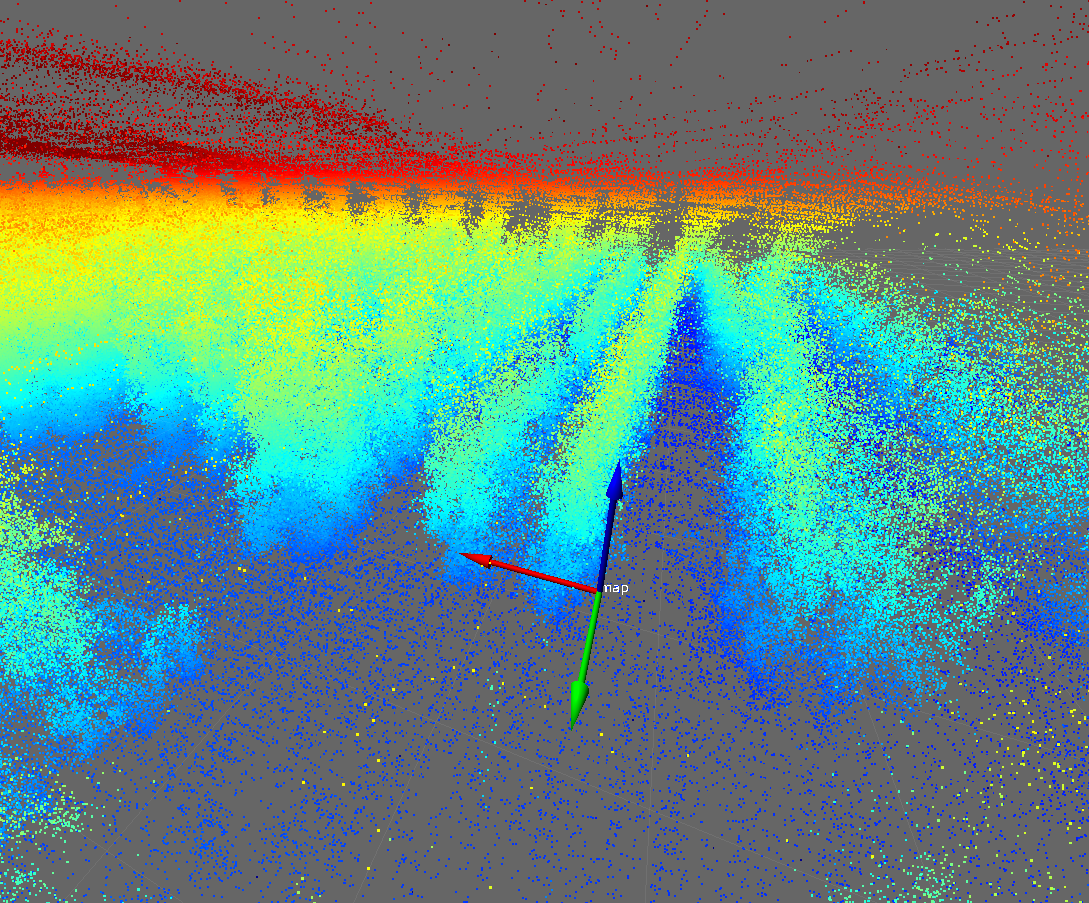}}%
    \label{subfig:2022-11-30}

    \caption{3D mapping of IFAPA greenhouse with Velodyne VLP16
    for sequeces 
    (a) \texttt{2022-10-14},
    (b) \texttt{2022-10-26},
    (c) \texttt{2022-11-09}.
    (d) \texttt{2022-11-30}.
    }
  \label{fig:SLAM_dias}
\end{figure}

Figure \ref{fig:SLAM_dias} shows the result of the greenhouse mapped with the Velodyne VLP16 LiDAR for four different days.

\begin{figure}[H]
  \centering
  \begin{subfigure}{0.45\textwidth}
    \centering
    \includegraphics[width=\textwidth]{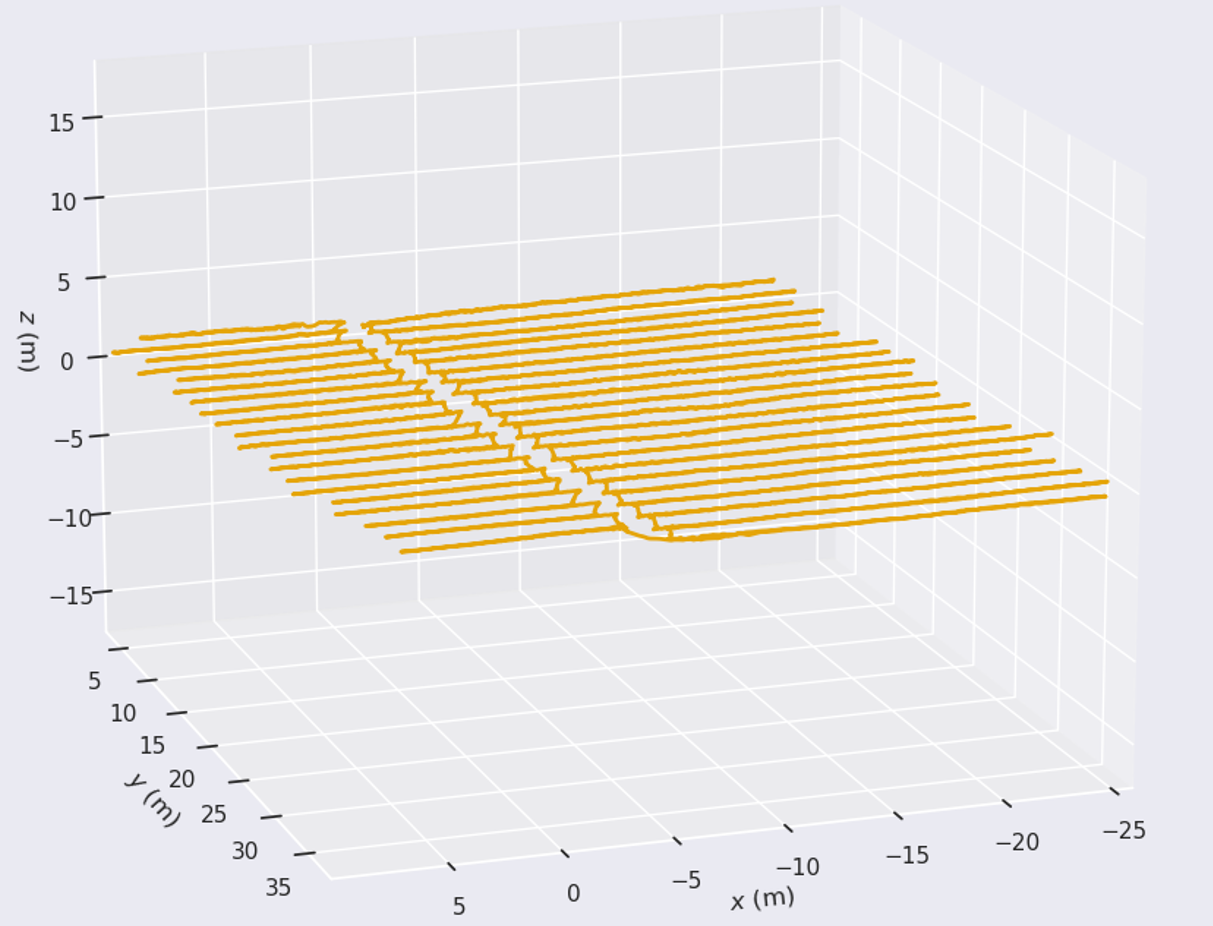}
    \caption{Velodyne tum path \texttt{2022-10-14}}
    \label{subfig:Path_2022-10-14}
  \end{subfigure}
  \hfill
  \begin{subfigure}{0.45\textwidth}
    \centering
    \includegraphics[width=\textwidth]{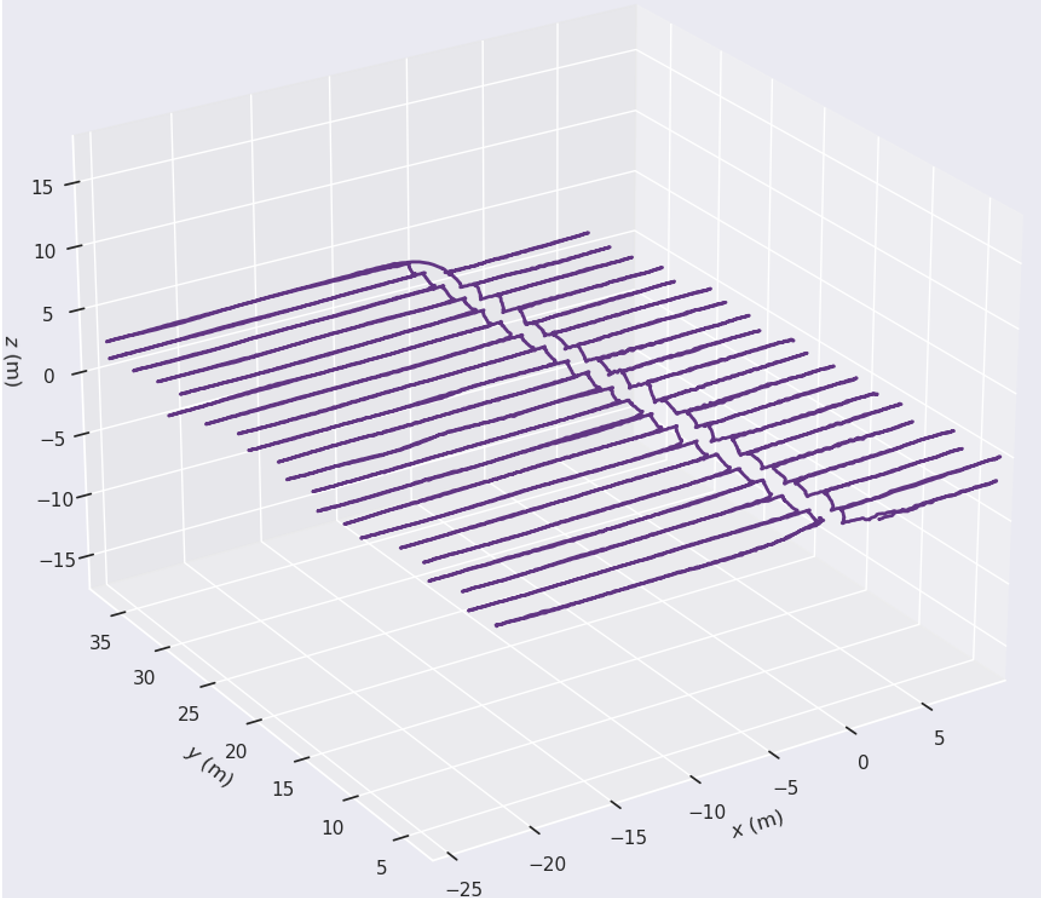}
    \caption{Velodyne tum path \texttt{2022-10-26}}
    \label{subfig:Path_2022-10-26}
  \end{subfigure}

  \vspace{1em}

  \begin{subfigure}{0.45\textwidth}
    \centering
    \includegraphics[width=\textwidth]{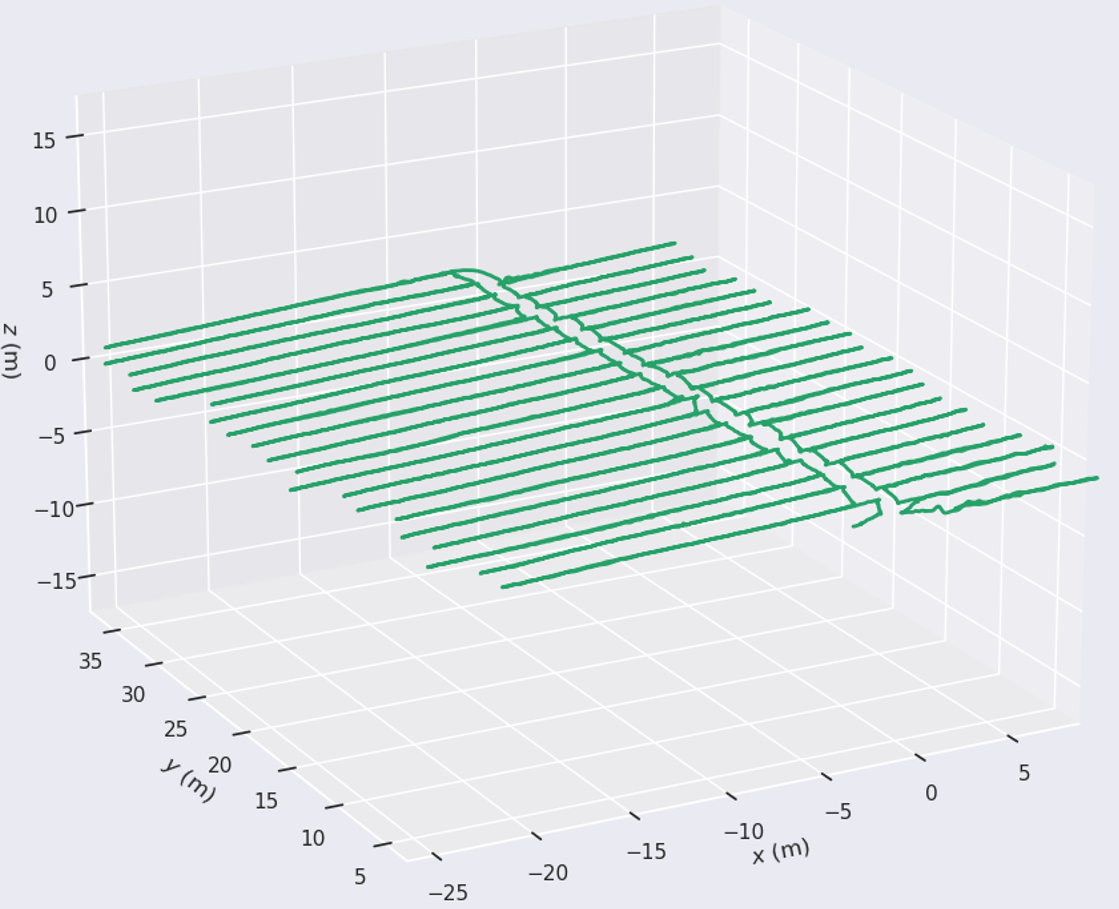}
    \caption{Velodyne tum path \texttt{2022-11-09}}
    \label{subfig:Path_2022-11-09}
  \end{subfigure}
  \hfill
  \begin{subfigure}{0.45\textwidth}
    \centering
    \includegraphics[width=\textwidth]{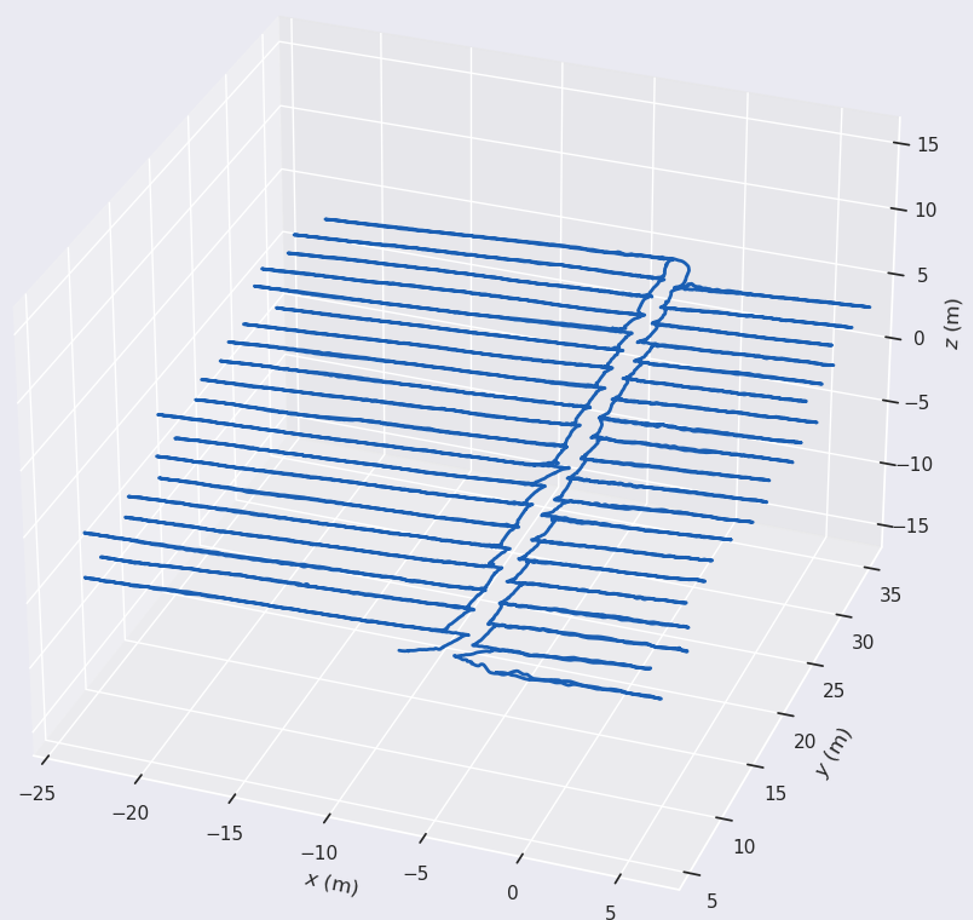}
    \caption{Velodyne tum path \texttt{2022-11-30}}
    \label{subfig:Path_2022-11-30}
  \end{subfigure}

  \caption{Path followed on different days with the Velodyne VLP16}
  \label{fig:Path_dias}
\end{figure}

A significant visual difference in the state of the tomato plants can be observed in Figures \ref{fig:SLAM_dias}(a)--(d), guaranteeing the systematic mapping of the greenhouse. At the top of each image,
phony LiDAR points are visible above the actual greenhouse
top surface.
These points are quite common for the VLP-16 LiDAR, which 
tends to detects phony points beyond the plastic surfaces of the greenhouse.
The greenhouse soil is often uneven, with slopes varying between 1 and 2 \%, depending on the area. In this type of mapping, being able to obtain a terrain variation model is of vital importance for robotics \cite{greenhouses20134}. 
As output of the SLAM module we also obtain the estimated platform trajectory (in the the standard \textit{TUM} format) which is then analyzed using \texttt{evo\_traj}
as shown in Figure~\ref{fig:Path_dias}.

\begin{figure}[H]
    \includegraphics[width=\linewidth]{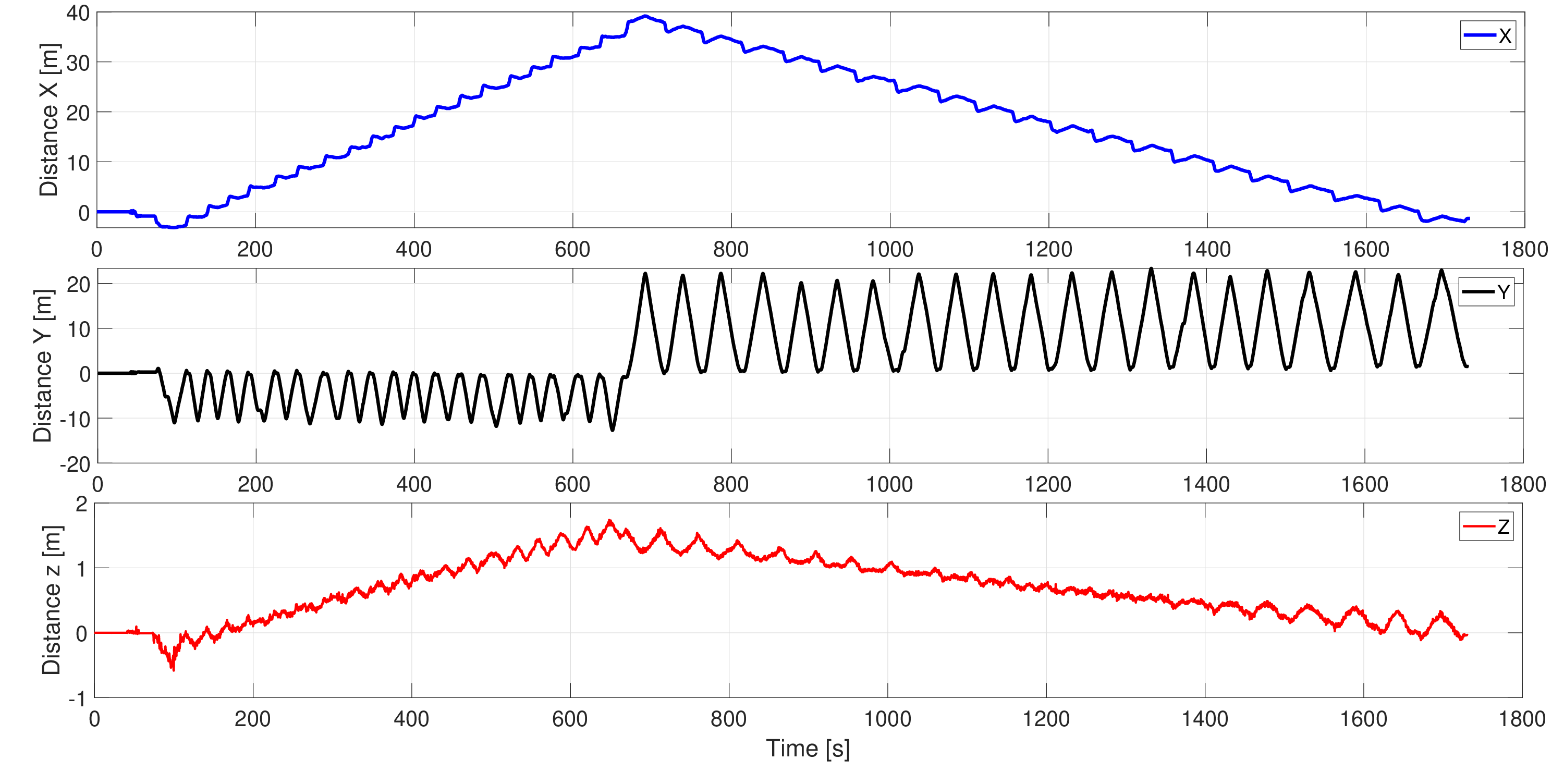} \centering
    \caption{Trajectory estimated from the Velodyne VLP16 data for sequence \texttt{2022\_11\_30}.}
    \label{fig:Z}
\end{figure}

It can be seen how the movement pattern is repeated every day. It can be seen how the pattern of movement is repeated every day. It is essential to mention that the trials start with the same orientation, as can be seen in Figures \ref{subfig:Trajectory_2022-10-14}, \ref{subfig:Trajectory_2022-10-26}, \ref{subfig:Trajectory_2022-11-09} and Figure \ref{subfig:Trajectory_2022-11-30} towards $+ x$ and $+ y$. Each axis behaves similarly irrespective of the day, although some exciting behaviors can be observed along each trajectory. To analyze this, the path followed by each axis on November 30th, 2022 is shown in Figure \ref{fig:Z}.

\begin{figure}[H]
    \includegraphics[width=\linewidth]{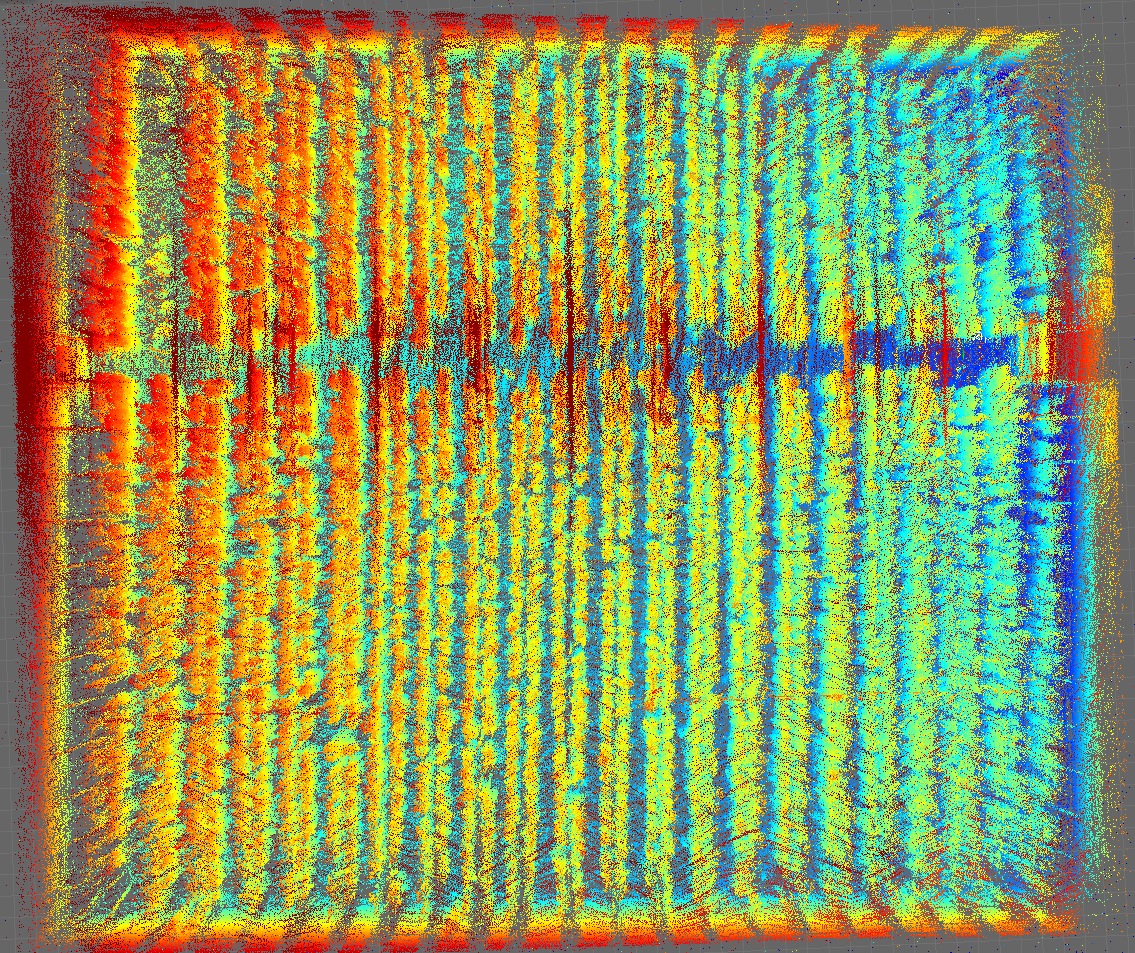} \centering
    \caption{Mapping of the complete greenhouse, plan view with Velodyne VLP16}
    \label{fig:Vel_SLAM3}
\end{figure}

The sub-graphs corresponding to the displacement in $x$ and $y$ correspond to the displacement in the 2D plane, but the graph in $z$ shows data relevant to the investigation. A noise is observed that comes from the deformed soil in the greenhouse, as well as a slope corresponding to the slope of the terrain itself, validated by the algorithm itself. Finally, Figures \ref{fig:Vel_SLAM2} and  \ref{fig:Vel_SLAM3} show a final orthogonal view and plan of the result of the mapped 3D model, observing the dataset's quality.

\subsection{Mapping - Ouster OS0}

\begin{figure}[H]
  \centering
  \begin{subfigure}{0.45\textwidth}
    \centering
    \includegraphics[width=\textwidth]{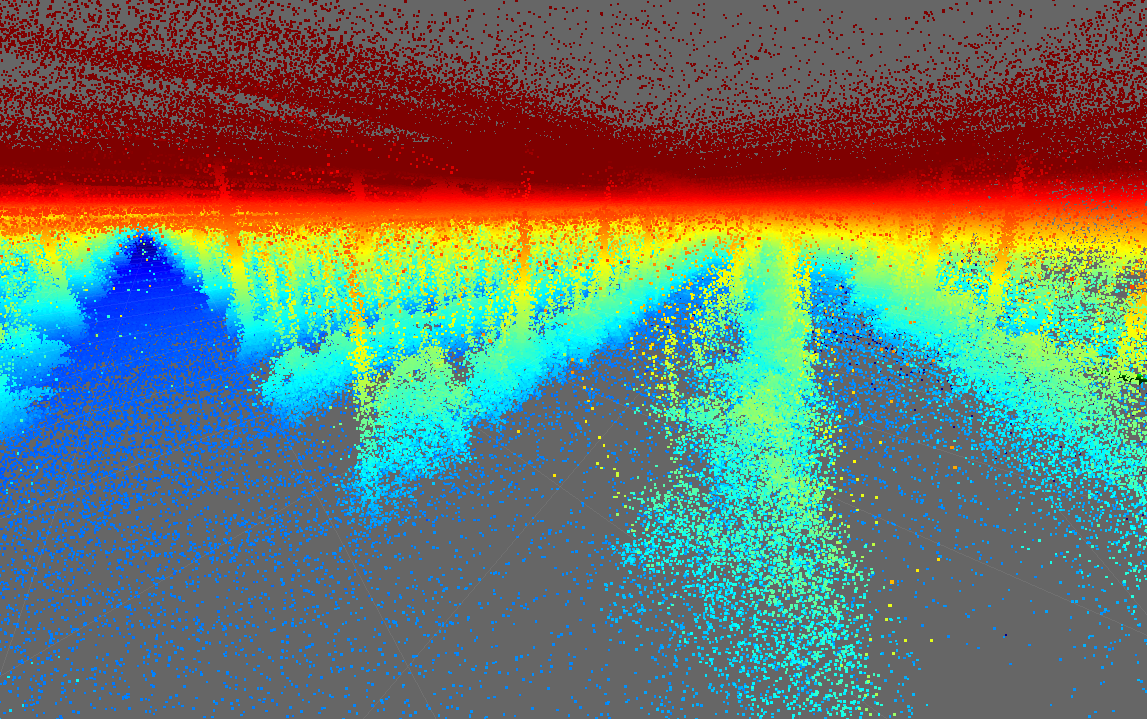}
    \caption{Ouster MOLA Lidar odometry \texttt{2022-10-14}}
    \label{subfig:O_2022-10-14}
  \end{subfigure}
  \hfill
  \begin{subfigure}{0.45\textwidth}
    \centering
    \includegraphics[width=\textwidth]{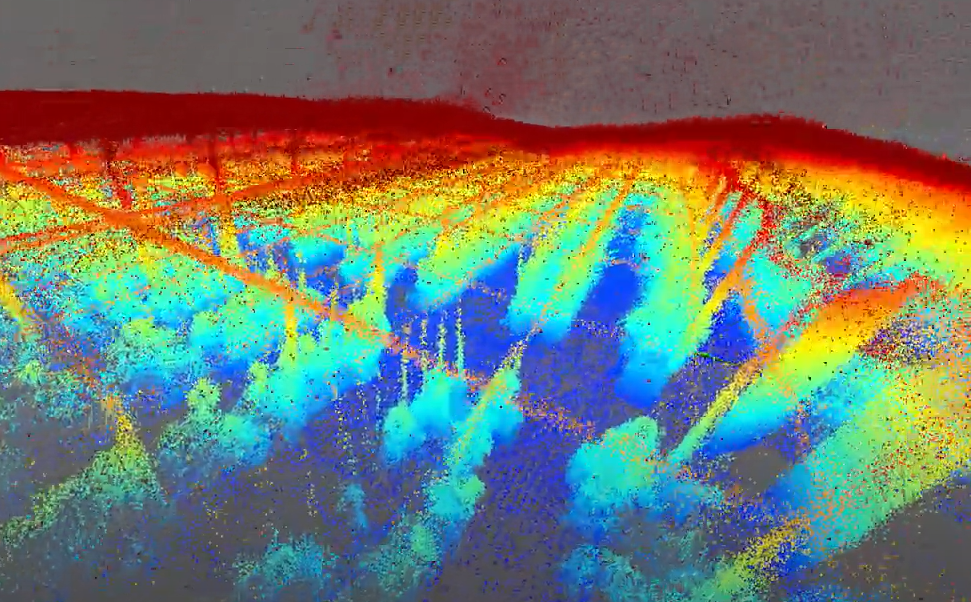}
    \caption{Ouster MOLA Lidar odometry \texttt{2022-10-26}}
    \label{subfig:O_2022-10-26}
  \end{subfigure}

  \vspace{1em}

  \begin{subfigure}{0.45\textwidth}
    \centering
    \includegraphics[width=\textwidth]{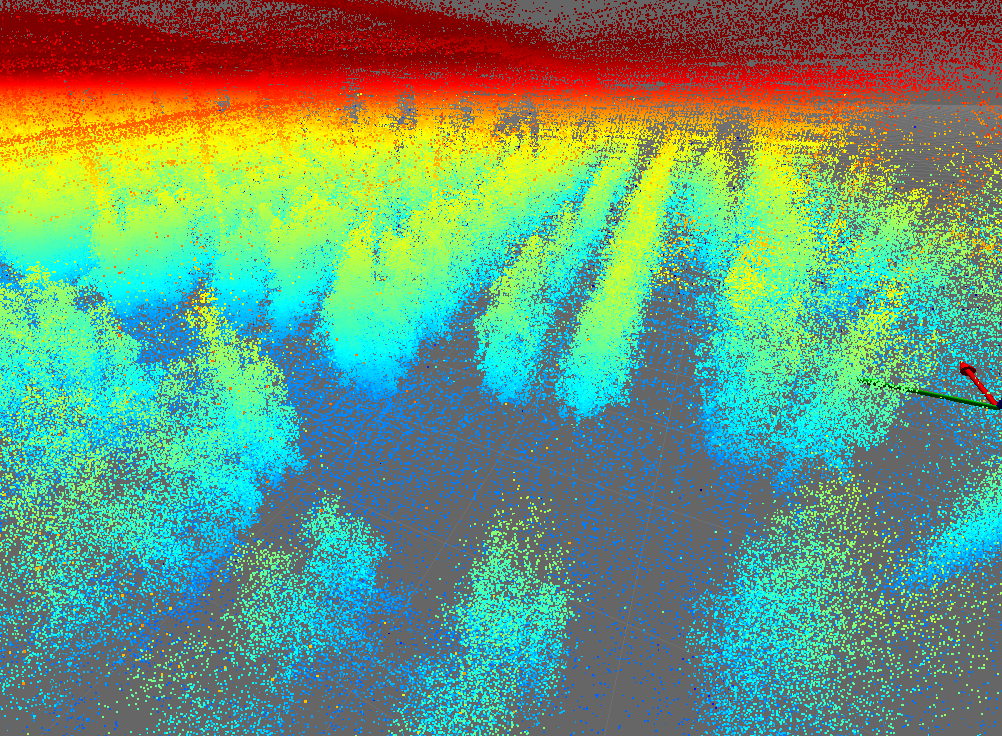}
    \caption{Ouster MOLA Lidar odometry \texttt{2022-11-09}}
    \label{subfig:O_2022-11-09}
  \end{subfigure}
  \hfill
  \begin{subfigure}{0.45\textwidth}
    \centering
    \includegraphics[width=\textwidth]{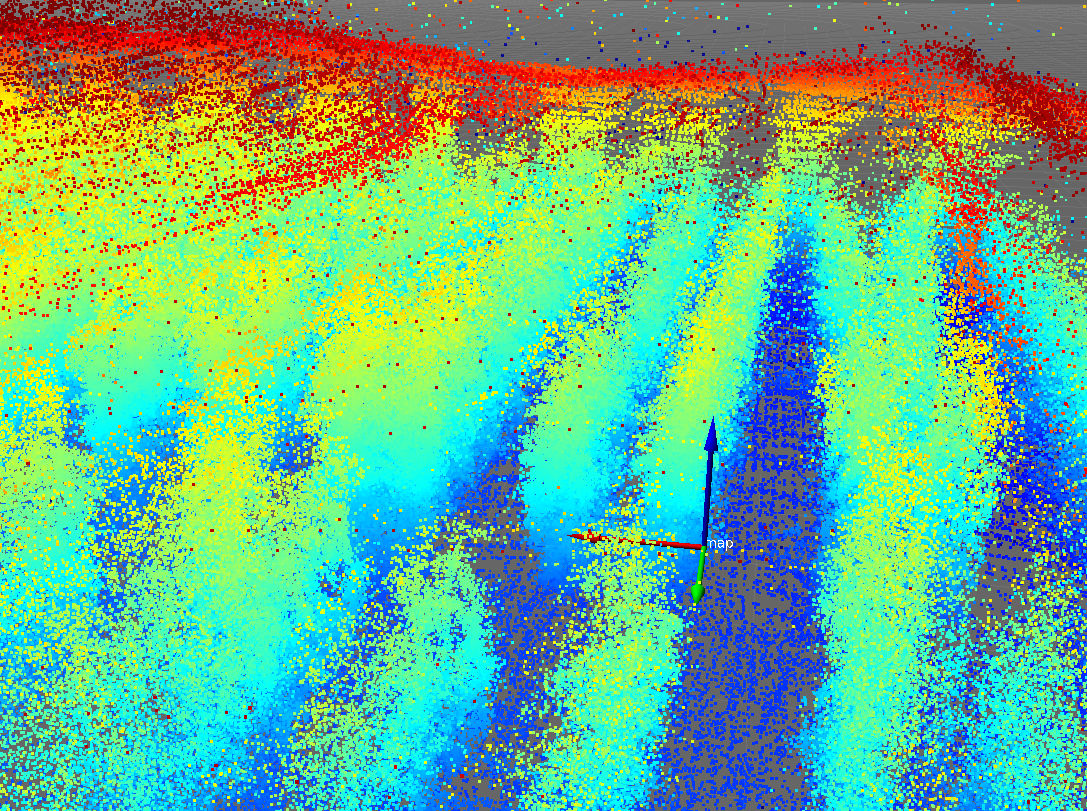}
    \caption{Ouster MOLA Lidar odometry \texttt{2022-11-30}}
    \label{subfig:O_2022-11-30}
  \end{subfigure}

  \caption{3D mapping of IFAPA greenhouse with Ouster OS0}
  \label{fig:O_SLAM_dias}
\end{figure}

\begin{figure}[H]
  \centering
  \begin{subfigure}{0.45\textwidth}
    \centering
    \includegraphics[width=\textwidth]{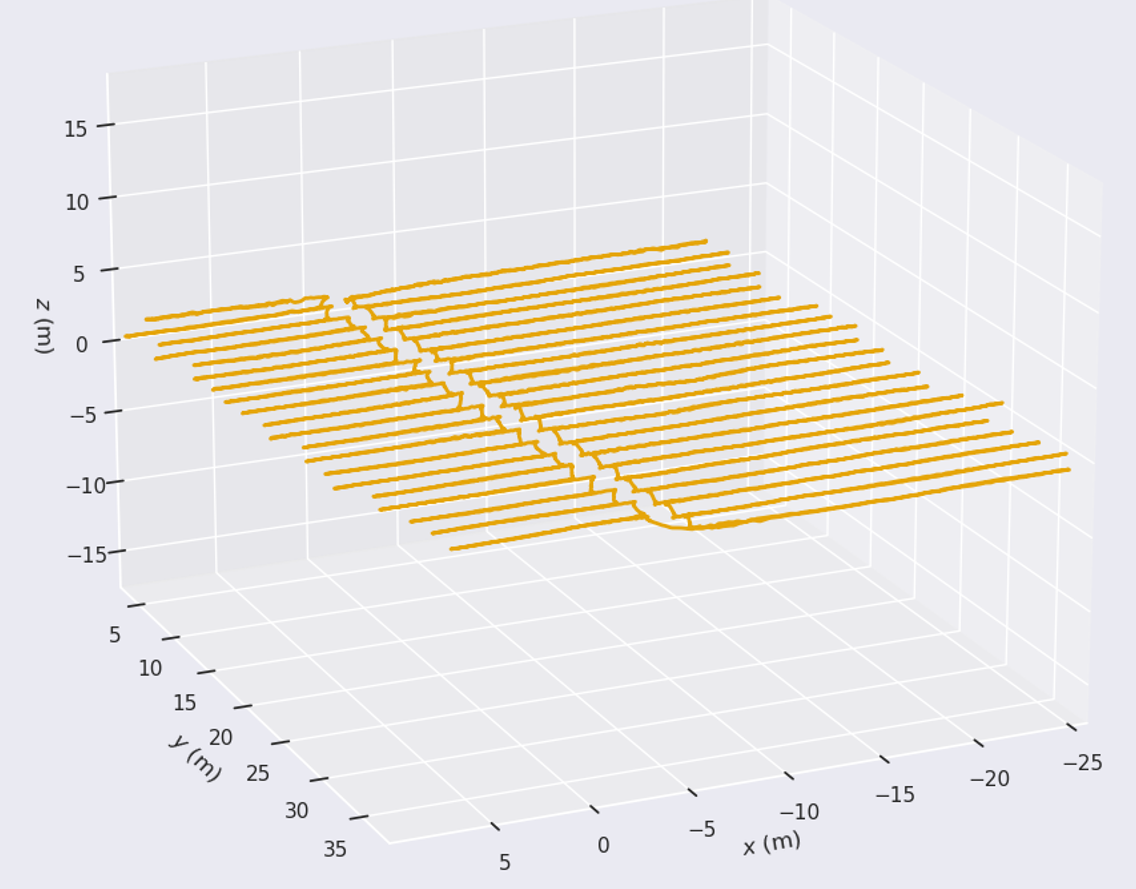}
    \caption{Ouster path \texttt{2022-10-14}}
    \label{subfig:O_Path_2022-10-14}
  \end{subfigure}
  \hfill
  \begin{subfigure}{0.45\textwidth}
    \centering
    \includegraphics[width=\textwidth]{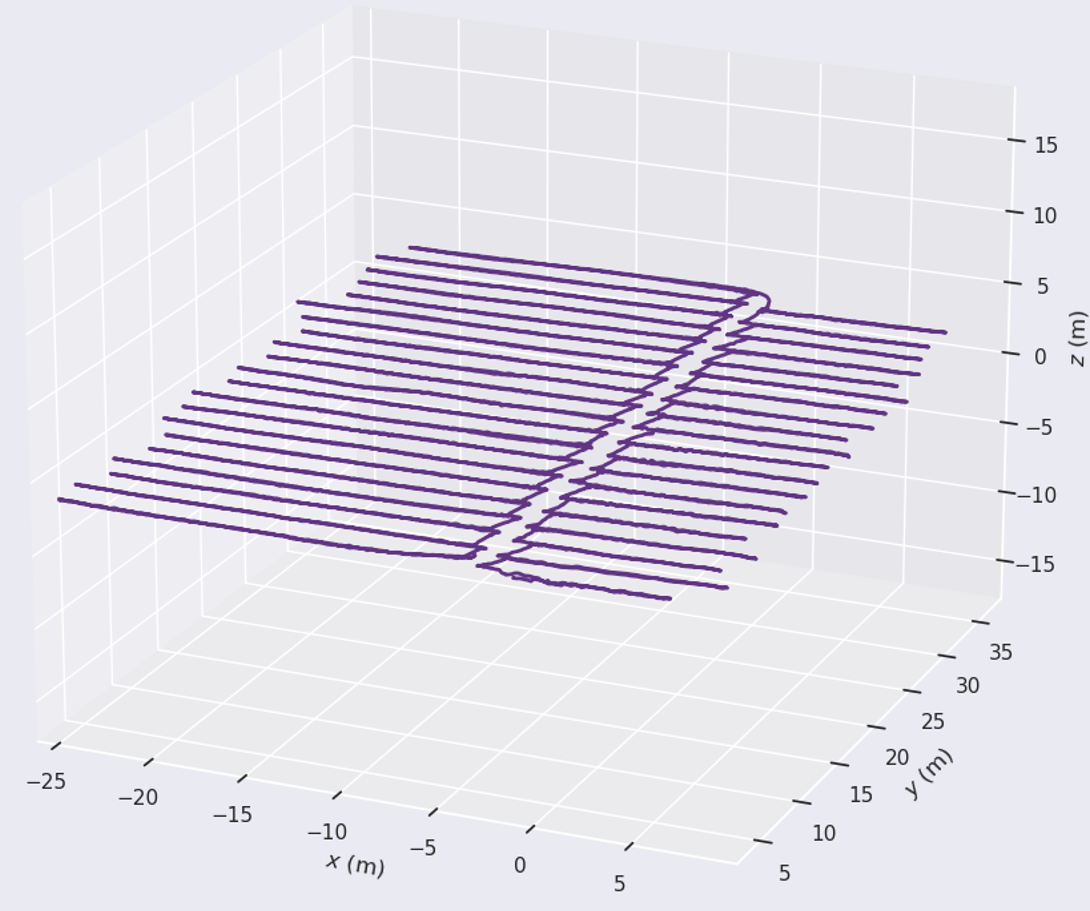}
    \caption{Ouster path \texttt{2022-10-26}}
    \label{subfig:O_Path_2022-10-26}
  \end{subfigure}

  \vspace{1em}

  \begin{subfigure}{0.45\textwidth}
    \centering
    \includegraphics[width=\textwidth]{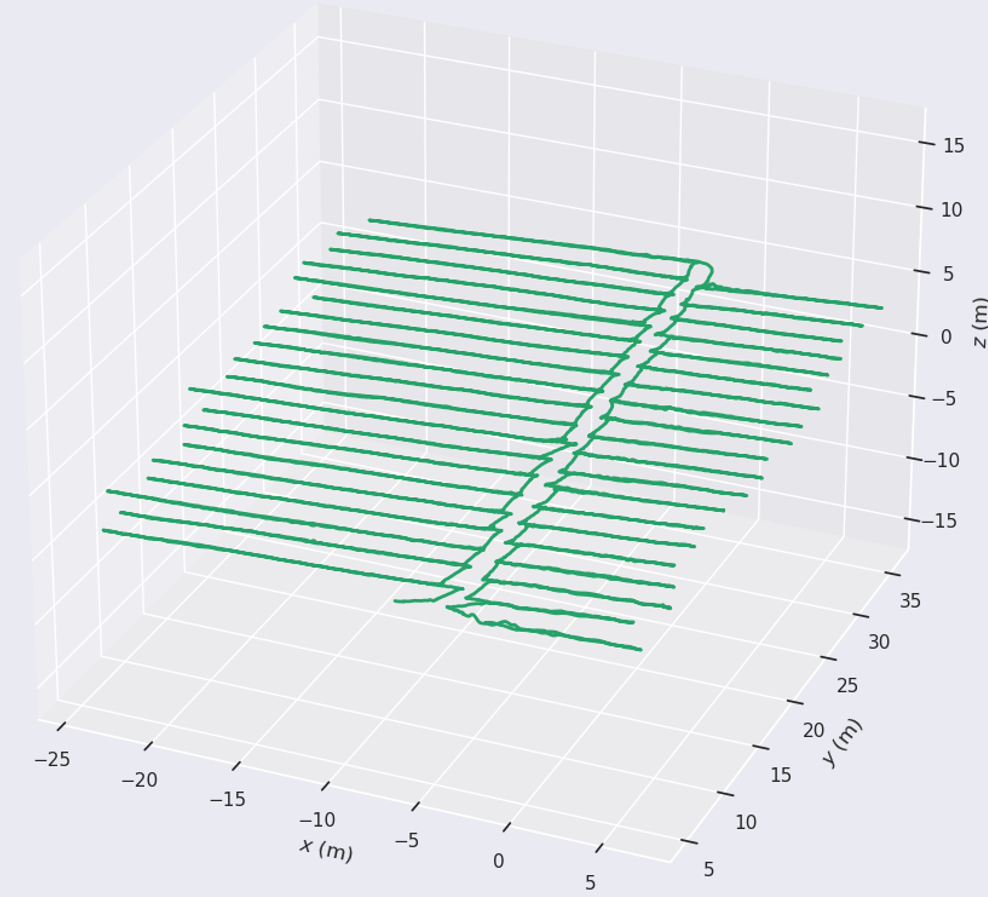}
    \caption{Ouster path \texttt{2022-11-09}}
    \label{subfig:O_Path_2022-11-09}
  \end{subfigure}
  \hfill
  \begin{subfigure}{0.45\textwidth}
    \centering
    \includegraphics[width=\textwidth]{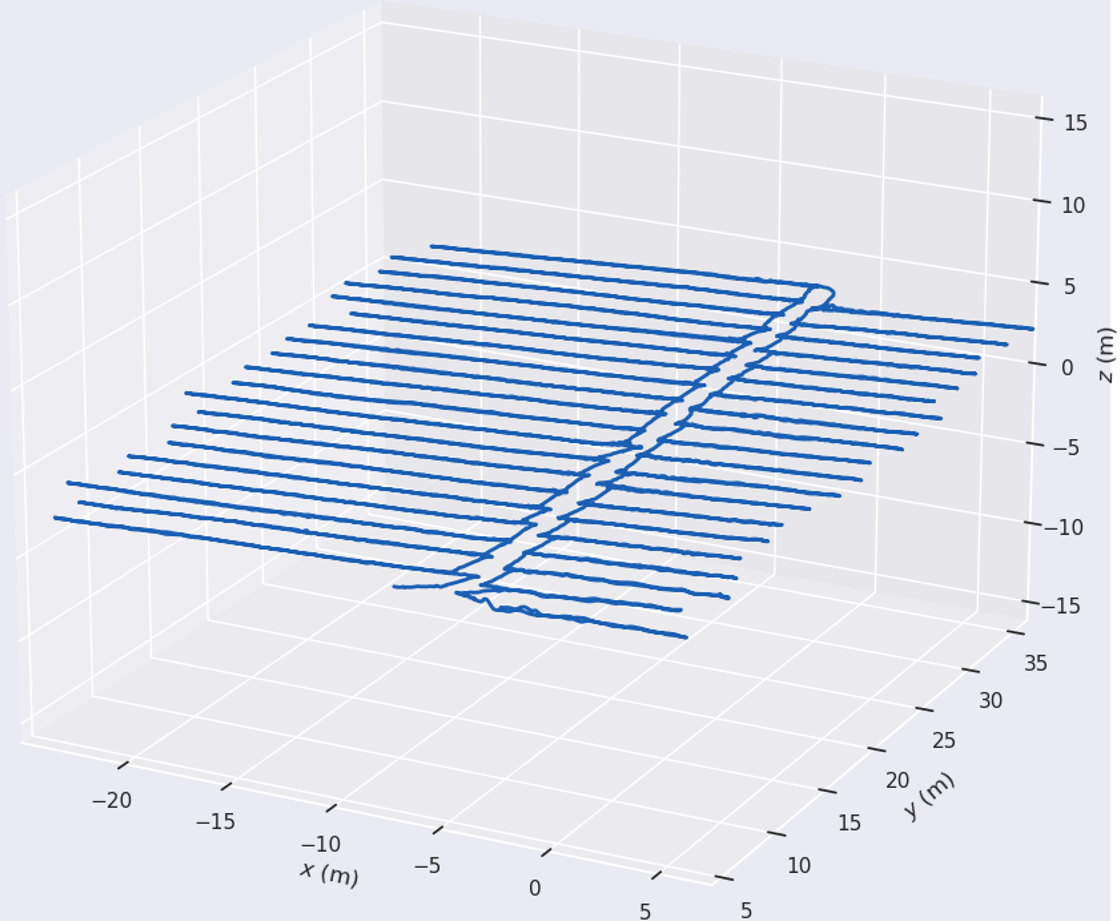}
    \caption{Ouster path \texttt{2022-11-30}}
    \label{subfig:O_Path_2022-11-30}
  \end{subfigure}

  \caption{Path followed on different days with the Ouster OS0}
  \label{fig:O_Path_dias}
\end{figure}

The same SLAM algorithm has been applied to the Ouster LiDAR data, obtaining the results shown in 
Figure~\ref{fig:O_SLAM_dias}.
In this case, the mapping is shown for the same days as analysed in the previous section. Similarly, Figure \ref{fig:O_Path_dias} shows the result of the trajectories recorded by the Ouster OS0.

A higher point cloud density can be observed as this LiDAR has twice point cloud rings than the VLP-16. 
In the same way, each of the segmented axes is projected, observing the same oscillatory behavior due to the uneven terrain.

\begin{figure}[H]
    \includegraphics[width=\linewidth]{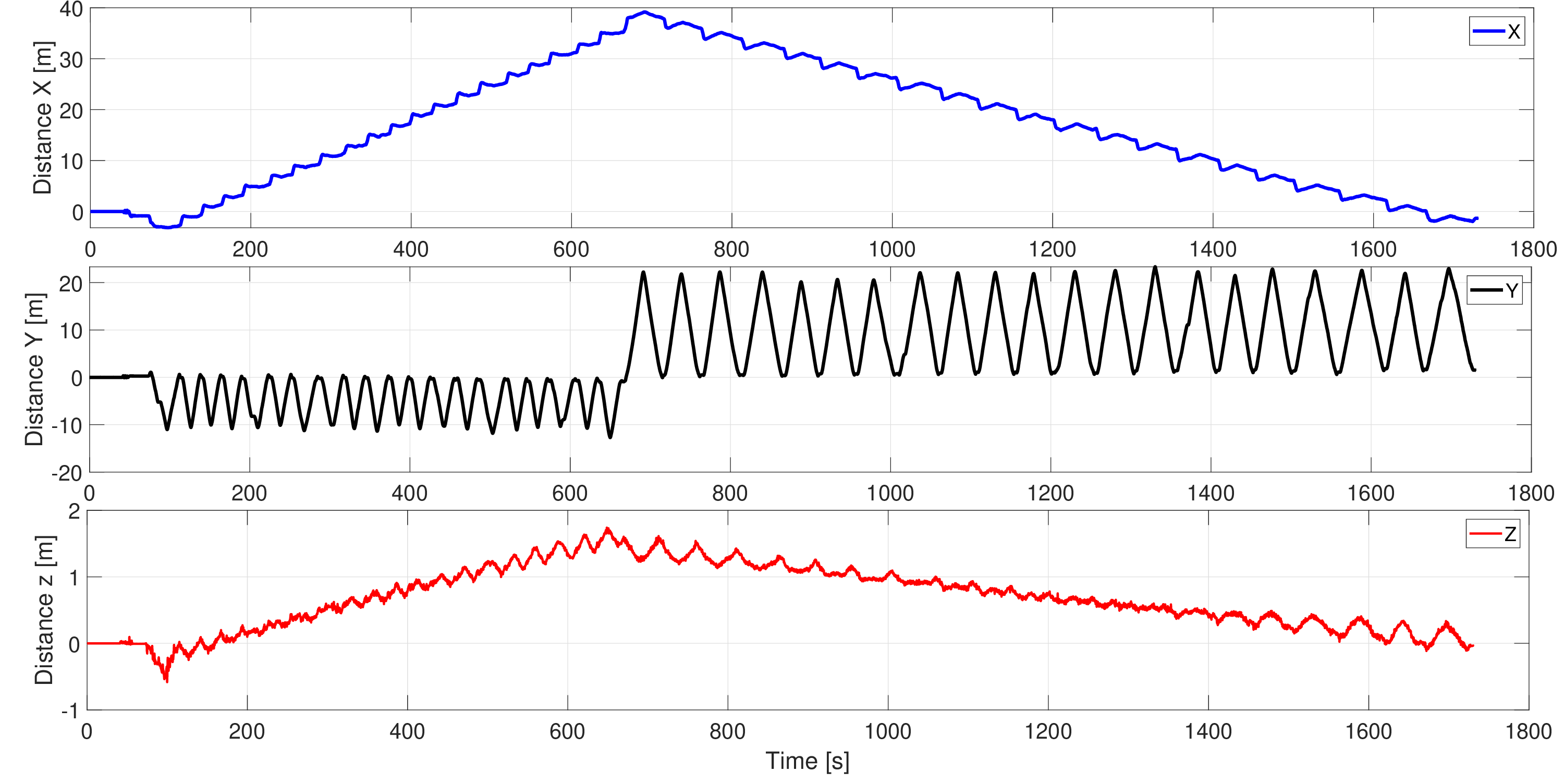} \centering
    \caption{Trajectory of each axis for sequence \texttt{2022\_11\_30} with Ouster OS0.}
    \label{fig:Z_O}
\end{figure}

\begin{figure}[H]
    \includegraphics[width=\linewidth]{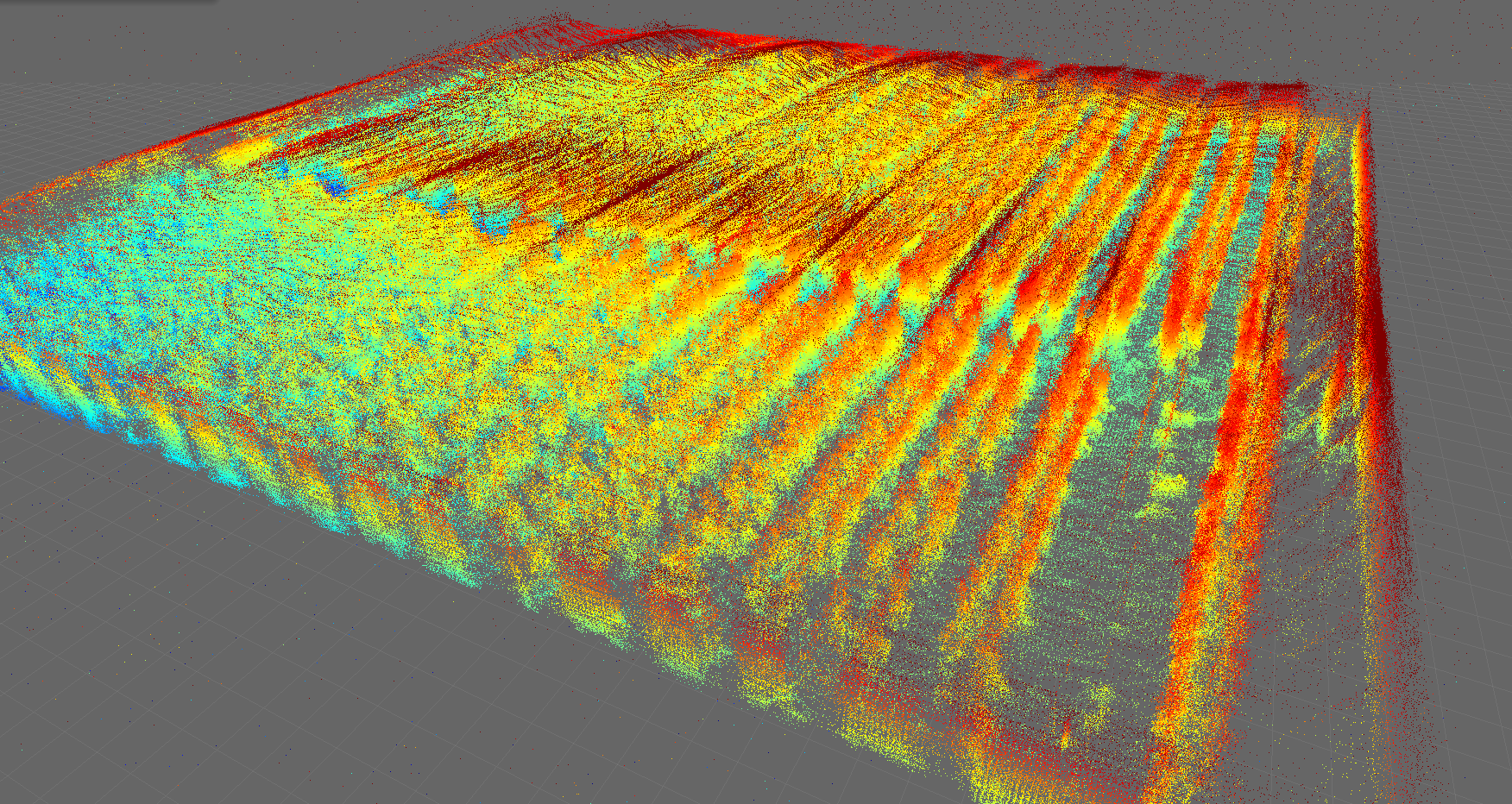} \centering
    \caption{Mapping of the complete greenhouse, orthogonal view with Ouster OS0}
    \label{fig:Ou_SLAM2}
\end{figure}

\begin{figure}[H]
    \includegraphics[width=\linewidth]{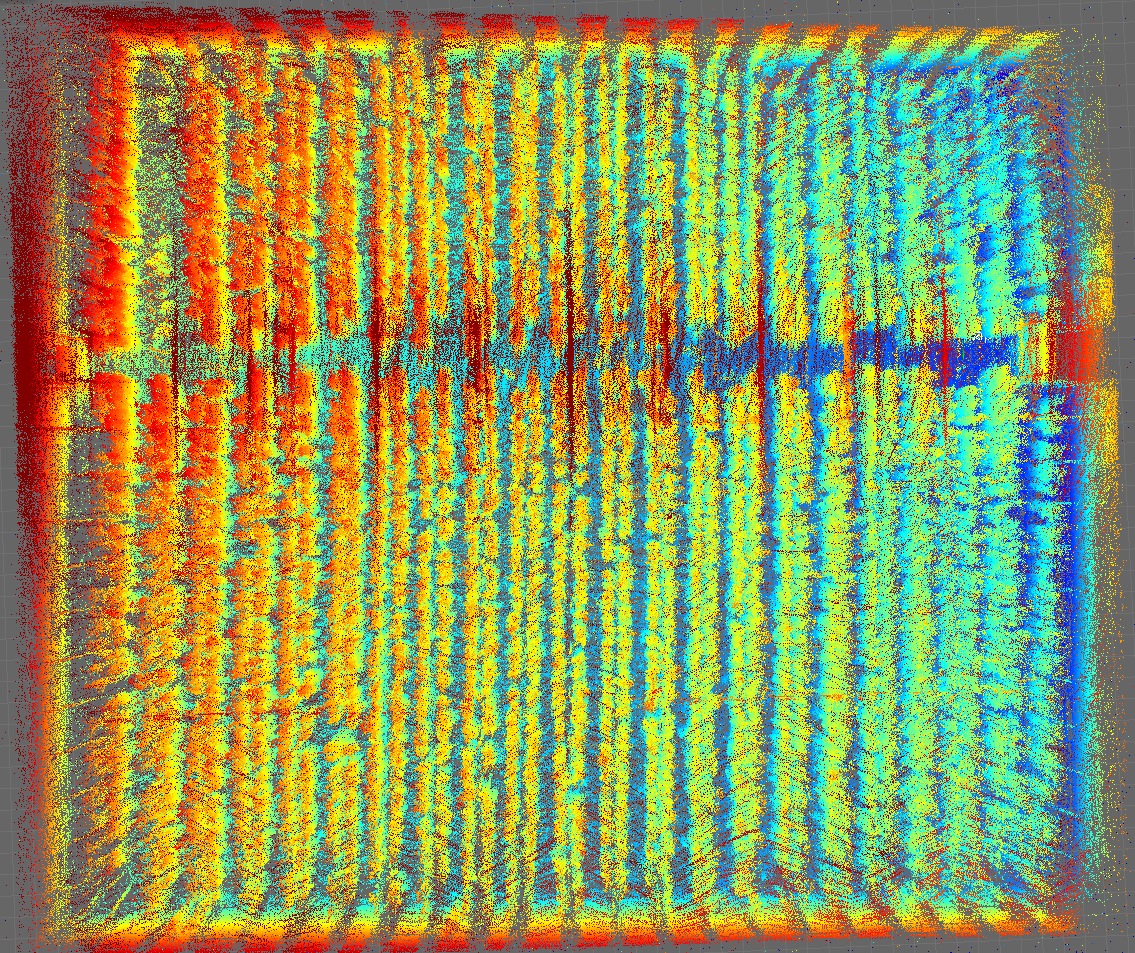} \centering
    \caption{Mapping of the complete greenhouse, plan view with Ouster OS0}
    \label{fig:ou_SLAM3}
\end{figure}

The original and plan views are presented in Figures \ref{fig:Ou_SLAM2} and \ref{fig:ou_SLAM3}, obtaining an excellent result and exceeding the dataset expectations.

\section{Conclusions and future work}

A new dataset designed for challenging natural agricultural environments is presented here to drive the development of autonomous robots that can improve operations in the agricultural sector. This article details the data collection and describes the process employed. In addition, the dataset is validated using a novel SLAM technique to ensure its operability. The results indicate that the data quality is suitable for implementation in specific algorithms. Providing open datasets, such as the one presented here, is critically important to further research and implementation of autonomous robots in agricultural applications in greenhouses, given the unique characteristics of these environments, which include extreme lighting variations, changing weather conditions, and seasonal cycles.

In particular, this dataset allows the Community to develop and test its own SLAM algorithms in a novel agricultural environment, with particular emphasis on intensive greenhouse cultivation. The ability to create a map and orientate and navigate with precision is essential for all tasks a robot can perform in a greenhouse (e.g., transport of vegetable crates, harvesting, spraying, data collection, etc).

The proposed framework is intended to be used as a database, which is unique for the time being. Data will continue to be collected from the greenhouse as all these sensors will be implemented in two robots of the University of Almería's own. It is important to note that it is intended to collect data more frequently with the implementation of robots, seeking a more extensive dataset.

Apart from installing these sensors in real robots, the possibility of installing RGB-D cameras on both sides of the robots will be explored to obtain detailed information about the crop. The implemented cameras will also be improved to make it more robust to sudden changes in lighting, crop growth, etc. All this will be implemented in ROS~2 Humble, providing the robot with the latest technology in mobile robotics. Finally, these models can be used for other tasks, so this exciting contribution could be used to obtain the semantic information of the crop type, determining a model of plant growth in height and width.

This paper aims to present and share a unique dataset with the potential to assist researchers in developing new SLAM-based techniques and in evaluating a totally novel environment such as a greenhouse. It is outside the scope of this work to compare and analyze the performance of existing methods. The results described in Section 5 are provided as reference that researchers can use as starting point for future comparative evaluations.

\section*{Funding}

This work has been partially financed by the 'CyberGreen' Project, PID2021-122560OB-I00, and the ‘AgroConnect.es’ infrastructure used to carry out this research, grant EQC2019-006658-P, both funded by MCIN/AEI/ 10.13039/501100011033 and by ERDF A way to make Europe. Also, author Fernando Cañadas-Aránega is supported by an FPI grant from Spanish Ministery of Science, Innovation and Universities.

\section*{Abbreviations}
The following abbreviations are used in this manuscript:

\begin{tabular}{@{}ll}
CUDA & Compute Unified Device Architecture\\
DDR4 &  Double Data Rate type four Synchronous Dynamic Random-Access Memory\\
GPS & Global Positioning System\\
Hum & Humidity\\
IMU & Inertial Measurement Unit\\
Ir & Irradiance\\
LiDAR & Light Detection And Ranging\\
MOLA & Modular Optimization framework for Localization and mApping\\
MRPT & Mobile Robot Programming Toolkit\\
RAM & Random Access Memory\\
RGB & Red, Green and Blue\\
RGB-D & Red Green Blue Depth (color + Depth channels)\\
ROS & Robot Operating System\\
SLAM & Simultaneous Localization And Mapping\\
Temp & temperature\\
TIFF & Tagged Image File Format\\
UAL & University of Almería\\
\end{tabular}

\bibliography{sensorbib}

\end{document}